\documentclass{article} % For LaTeX2e
\makeatletter
\@ifpackageloaded{xcolor}
  {\PassOptionsToPackage{table}{xcolor}}
  {\usepackage[table]{xcolor}}
\makeatother
\usepackage{iclr2026_conference,times}

\usepackage{amsmath,amsfonts,bm}

\def\eqref#1{equation~\ref{#1}}
\def\1{\bm{1}}

\DeclareMathAlphabet{\mathsfit}{\encodingdefault}{\sfdefault}{m}{sl}
\SetMathAlphabet{\mathsfit}{bold}{\encodingdefault}{\sfdefault}{bx}{n}

\DeclareMathOperator{\Tr}{Tr}

\usepackage{wrapfig}
\usepackage[font=small]{caption}
\usepackage{hyperref}
\hypersetup{hidelinks}
\usepackage{url}
\usepackage{booktabs}
\usepackage{amsmath, amssymb, amsthm, bm}
\usepackage{multirow}
\usepackage{mathtools}
\usepackage{algorithm}

\usepackage{algorithmicx}
\usepackage{graphicx}
\usepackage{adjustbox}
\usepackage{subcaption}
\usepackage{algpseudocode}
\usepackage{soul}
\usepackage{wrapfig}
\usepackage{breqn}
\usepackage{float}
\newcommand{\conf}[1]{$_{\mathrm{#1}}$}
\DeclareMathOperator{\stopgrad}{stopgrad}

\newcommand{\Dt}{\mathcal{D}}   % dataset
\newcommand{\cmark}{\checkmark}
\newcommand{\xmark}{\(\times\)}
\newcommand{\EE}{\mathbb{E}}

\theoremstyle{plain}

\theoremstyle{remark}

\theoremstyle{definition}

\title{Two-Way Is Better Than One:
Bidirectional Alignment with Cycle Consistency for Ex\-emplar-Free Class-Incremental Learning}
\author{Hongye Xu \\
Chester F. Carlson Center for Imaging Science \\
Rochester Institute of Technology \\
\texttt{hx5239@rit.edu}
\And
Bartosz Krawczyk \\
Chester F. Carlson Center for Imaging Science \\
Rochester Institute of Technology \\
\texttt{bartosz.krawczyk@rit.edu}
}
\iclrfinalcopy % Uncomment for camera-ready version, but NOT for submission.
\begin{document}
\lhead{Published as a conference paper at ICLR 2026}

\maketitle
\begin{abstract}
Continual learning (CL) seeks models that acquire new skills without erasing prior knowledge. In exemplar-free class-incremental learning (EFCIL), this challenge is amplified because past data cannot be stored, making representation drift for old classes particularly harmful. Prototype-based EFCIL is attractive for its efficiency, yet prototypes drift as the embedding space evolves; thus, projection-based drift compensation has become a popular remedy. We show, however, that existing one-directional projections introduce systematic bias: they either retroactively distort the current feature geometry or align past classes only locally, leaving cycle inconsistencies that accumulate across tasks. We introduce \textbf{BiCyc}, a \textbf{bi}directional projector alignment approach with a \textbf{cyc}le-consistency objective: two maps, old$\to$new and new$\to$old, are optimized with stop-gradient gating so that transport and representation co-evolve. Analytically, we prove that the cycle loss contracts the singular spectrum toward unity in whitened space and that improved transport of class means/covariances yields smaller perturbations of classification log-odds, preserving old-class decisions and directly mitigating catastrophic forgetting. Empirically, across standard EFCIL benchmarks, our method substantially reduces forgetting and improves accuracy in from-scratch settings, while remaining competitive in the pretrained fine-grained regime.
The code is available at \url{https://github.com/HXuSz11/BiCyc_ICLR2026}.
\end{abstract}
\vspace{-2mm}
\section{Introduction}
\vspace{-2mm}
\label{intro}

Continual learning (CL) studies models that learn from a stream of tasks without retraining from scratch or erasing prior knowledge~\citep{Parisi:2019,Lange:2022,zenke2017continual}. A widely used protocol is \emph{class-incremental learning} (CIL), where tasks introduce disjoint labels and the learner must recognize all seen classes at test time without task identifiers. While rehearsal with stored exemplars often curbs forgetting~\citep{lopez2017gradient,riemer2018learning,pham2021dualnet,caccia2021new,wang2022meta,yang2023efficient}, privacy or memory constraints motivate the \emph{exemplar-free} variant (EFCIL), which prohibits retaining raw inputs. Among the many directions to mitigate forgetting~\citep{zenke2017continual,lopez2017gradient,schwarz2018progress,aljundi2018memory,riemer2018learning,serra2018overcoming,saha2020gradient,pham2021dualnet,caccia2021new,deng2021flattening,cha2021co2l,wang2022coscl,wang2022meta,slim2022_transil,wang2023dualhsic,yang2023efficient,UDIL,wang2024a}, prototype-based EFCIL has emerged as a compelling compromise: the model caches compact class statistics (means/covariances), and inference proceeds via nearest-prototype or Bayes scores—achieving strict no-memory operation with modest compute.

The core difficulty in prototype-based EFCIL is representation drift: as the backbone adapts to new tasks, the embedding geometry shifts and previously cached statistics become stale, biasing predictions toward recent classes. Existing EFCIL solutions to drift largely follow two routes that differ in how they balance stability and plasticity.

\textbf{Covariance and geometry modeling.}
This route improves robustness by shaping the feature geometry or the decision metric, commonly keeping the backbone partially/fixed to limit drift. FeTrIL~\citep{petit2023fetril} freezes the backbone and translates features to synthesize pseudo-features for past classes, trading some plasticity for stability. FeCAM~\citep{toldo2022fecam} argues that Euclidean metrics are suboptimal under non-stationarity and adopts anisotropic (Mahalanobis) scoring with class-wise covariances, typically with a frozen extractor. PASS~\citep{zhu2021pass} strengthens old-class representations via prototype augmentation and self-supervision without exemplars. These methods effectively mitigate forgetting by stabilizing or re-weighting the geometry, but they largely \emph{avoid cross-space transport}; the price of stability is potentially limited adaptation to new tasks.

\begin{wrapfigure}[17]{r}{0.40\columnwidth} 
  \vspace{0pt} 
  \centering
  \includegraphics[width=\linewidth]{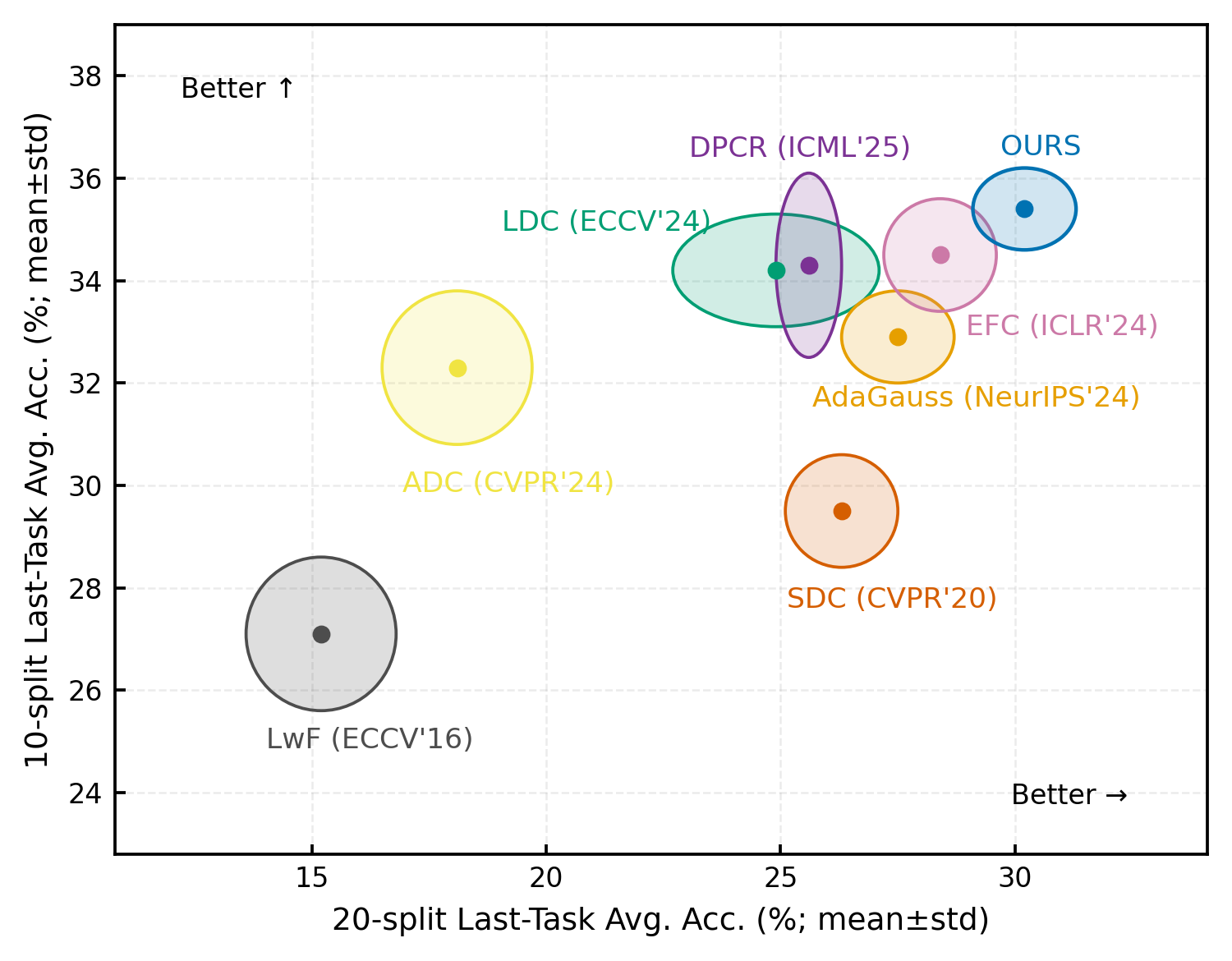}
  \caption{TinyImageNet ($T{=}10$): Our training algorithm yields solid
performance gains over state-of-the-art EFCIL methods.}
  \label{fig:acc_vs_forget}
  \vspace{-6pt}
\end{wrapfigure}

% \begin{wrapfigure}[20]{r}{0.40\columnwidth} 
%   \vspace{0pt} 
%   \centering
%   \includegraphics[width=\linewidth]{figs/acc_vs_forget_7methods.pdf}
%   \caption{CIFAR-100 ($T{=}10$): Our training algorithm yields solid
% performance gains over state-of-the-art EFCIL methods.}
%   \label{fig:acc_vs_forget}
%   \vspace{-6pt}
% \end{wrapfigure}

\textbf{Prototype drift compensation.}
A second—and increasingly dominant—route retains backbone plasticity and explicitly \emph{transports} outdated prototypes into the current space. SDC~\citep{0004TLHWCJ020} projects new features toward the old space and updates old prototypes accordingly. ADC~\citep{GoswamiSLKT024} estimates drift adversarially by pushing new samples toward old prototypes, then “resurrects’’ past classes. LDC~\citep{GomezVillaGWBTW24} replaces hand-crafted updates with a learnable drift module that scales across regimes. EFC~\citep{Magistri2024EFC} performs affinity-weighted, class-wise shifts that update prototypes in tandem with classifier training. AdaGauss~\citep{rypesc2024task} follows the learned-projector path but transports full Gaussian class statistics (means and covariances) into the new space for Bayesian inference rather than only moving class means. DPCR~\citep{hesemantic} mitigates representation drift by estimating task-to-task semantic shift from stored class statistics and then analytically re-calibrating the classifier via ridge regression using only these cached moments, avoiding any exemplar replay. Despite strong performance, the prevailing paradigm here is \emph{two-stage}: first train on the new task (often with regularization/distillation), then learn a post-hoc adapter (old$\!\to\!$new). This paradigm leaves residual inconsistencies between spaces: transport is optimized only after the fact, and cycle errors accumulate over tasks.
% \textcolor{blue}{DPCR~\citep{hesemantic} combines shift estimation with a closed-form classifier reconstruction that rebalances old and new classes from updated prototypes and covariances.} 

\textbf{Our idea: from two-stage to \emph{near} single-stage transport.}
Motivated by the limitations of two-stage drift compensation, we propose bidirectional cycle consistency that evolves adapter training \emph{into} the main task optimization so that transport and representation co-evolve. Concretely, during each new task we jointly learn two maps—\(A: z_{\text{old}}\!\to\! z_{\text{new}}\) and \(D: z_{\text{new}}\!\to\! z_{\text{old}}\)—with \emph{stop-gradient} targets to prevent retrograde updates on the evolving representation and a \emph{cycle-consistency} objective that regularizes the pair toward a near-bijection on the data support. Analytically, we show that the cycle loss in whitened space equals $\|\tilde A\tilde D-I\|_F^2$ and contracts the singular spectrum of $\tilde A\tilde D$ toward one; and that smaller alignment/cycle errors yield tighter bounds on the perturbation of classification log-odds, preserving old-class decisions. After the main stage, a brief consolidation fine-tune is applied; inference uses a Gaussian Bayes classifier built from transported old-class statistics and freshly estimated current-task statistics.

\noindent\textbf{Contributions.}
\begin{itemize}
\item \textbf{Bidirectional cycle consistency (BiCyc) within training.} We formulate paired projections \(A\) (old $\!\to\!$ new) and \(D\) (new $\!\to\!$ old) learned \emph{during} the task, with stop-gradient gating and a cycle loss that enforces near-inverse behavior on-support—addressing the asymmetry and post-hoc mismatch of prior two-stage, one-way pipelines.
\item \textbf{Geometry-preserving transport for drift mitigation.} Our transport keeps old-class geometry stable as the backbone changes, yielding reduced recency bias and higher knowledge retention.
\item \textbf{Theory-grounded alignment.} We prove that minimizing the cycle loss contracts the singular spectrum toward unity in whitened space and derive bounds linking mean/covariance transport errors to classification log-odds stability, explaining the observed reduction in forgetting.
% \item \textbf{Seamless prototype transport for balancing stability and plasticity.} Post training, we evaluate with a Gaussian Bayes classifier that combines transported old-class means/covariances (via $A$) with current-task estimates, yielding a unified, exemplar-free decision rule.
\item \textbf{Near single-stage pipeline with strong results.} By collapsing adapter learning into the main stage (with a short consolidation fine-tune), our method strikes an excellent balance between preserving stability (i.e., preventing drift) and maintaining plasticity, substantially reducing forgetting and maintaining or improving new-task accuracy across CIFAR-100, TinyImageNet, ImageNet-100, and CUB-200 under multiple splits. We discuss limitations in the experiments section.
\end{itemize}

\begin{figure}[t]
  \centering
  \begin{minipage}{\columnwidth}
    % \captionsetup{width=\linewidth, justification=raggedright,
    %               singlelinecheck=false}

    \includegraphics[width=\linewidth]{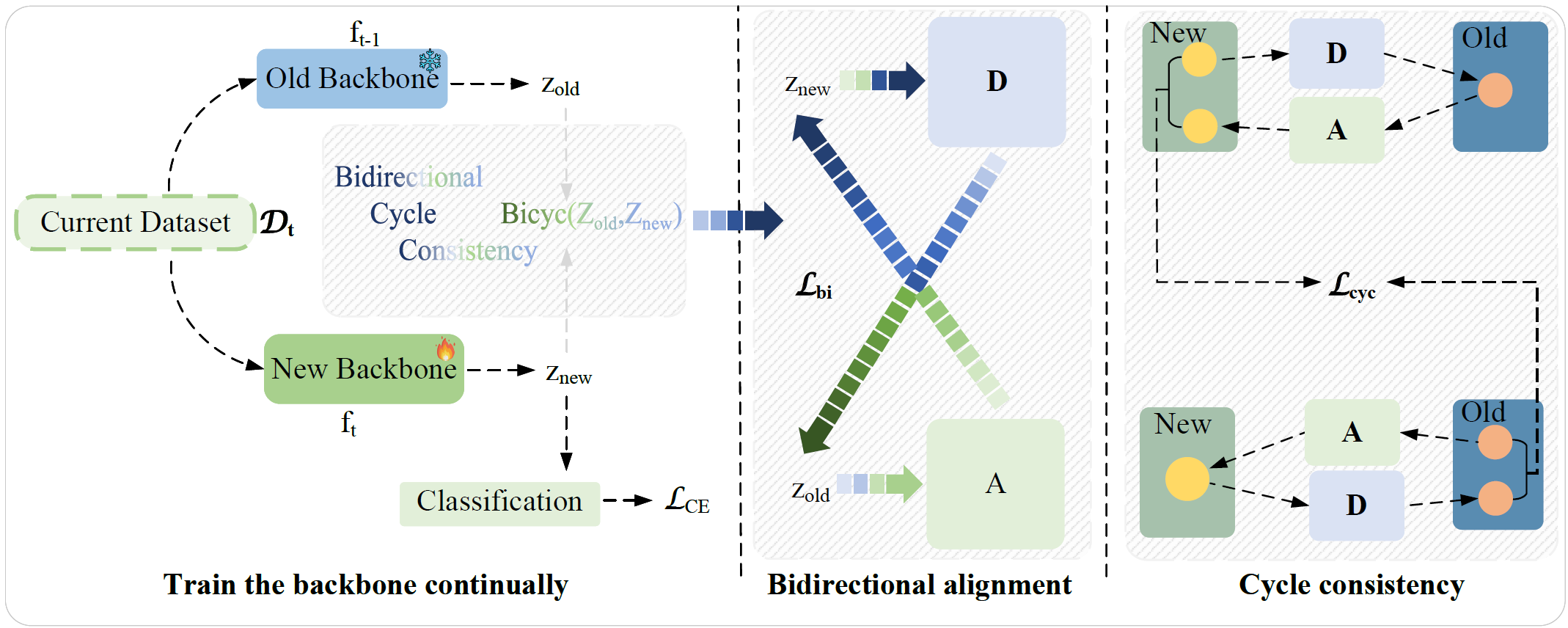}
    \caption{\textbf{Overview.} \textbf{(1) Train:} the current backbone $f_t$
    learns on $\mathcal{D}_t$ (producing $z_{\text{new}}$, while frozen $f_{t-1}$
    provides $z_{\text{old}}$) with task loss $\mathcal{L}_{\mathrm{CE}}$.
    \textbf{(2) Bidirectional alignment:} jointly learn a distiller
    $D\!:z_{\text{new}}\!\to\!z_{\text{old}}$ and an adapter
    $A\!:z_{\text{old}}\!\to\!z_{\text{new}}$ using $\mathcal{L}_{\mathrm{bi}}$.
    \textbf{(3) Cycle consistency:} enforce $A\!\circ\!D\!\approx\!I$ and
    $D\!\circ\!A\!\approx\!I$ with $\mathcal{L}_{\mathrm{cyc}}$,
    yielding near-bijective, geometry-preserving transport. Old Gaussian prototypes
    are mapped forward by $A$, and all classes are evaluated in the \emph{new} space.}
    \vspace{-4mm}
    \label{fig:main_pipline}
  \end{minipage}
\end{figure}

\vspace{-2mm}
\section{Preliminaries}
\vspace{-2mm}
\subsection{Problem Definition}
\vspace{-1mm}
\label{sec:problem_def}

Continual learning (CL) aims to train a model on a stream of tasks while preserving previously acquired knowledge.
In the \textbf{class-incremental} scenario considered here, each task
$t \in \{1,\dots,T\}$ introduces a disjoint label set $\mathcal{C}_t$
with $\mathcal{C}_i \cap \mathcal{C}_j = \varnothing$ for $i \neq j$.
After learning task~$t$, the model must recognize any class in
$\mathcal{C}_{1:t} := \bigcup_{i=1}^{t} \mathcal{C}_i$
without a task identifier at test time.

Let $f_t : \mathcal{X} \to \mathbb{R}^d$ denote the feature extractor after completing the first $t$ tasks.
During training on task~$t$, the learner has access only to
$\mathcal{D}_t = \{(x_i, y_i) \mid y_i \in \mathcal{C}_t\}$. % ★ 统一 $D_t \to \mathcal{D}_t$
The absence of any prior-task data defines the exemplar-free class-incremental setting.

\vspace{-1mm}
\subsection{Prototype-Based Exemplar-Free CIL}
\vspace{-1mm}
\label{sec:efcil}

In \textbf{exemplar-free class-incremental learning (EFCIL)}, the learner is prohibited from retaining raw samples from prior tasks. In the absence of replayed data, a common strategy is to summarize past
knowledge with \emph{prototypes}—one representative feature mean per seen class.
Focusing on a single transition $t{-}1\!\to\!t$, after completing task $t{-}1$
the learner stores for each class $c\in\mathcal{C}_{1:t-1}$ the prototype.
\begin{equation}
\label{eq:proto_def}
\bm{\mu}_c^{\,t-1}
=\frac{1}{|\mathcal{D}_c|}
\sum_{x\in\mathcal{D}_c} f_{t-1}(x),
\end{equation}
where $\mathcal{D}_c$ collects all examples of class $c$ encountered up to
step $t{-}1$. This summary is compact—its memory scales as
$\mathcal{O}(|\mathcal{C}_{1:t-1}|\,d)$ for feature dimension $d$—and can
be used at inference time either directly with a nearest-class-mean rule
or to regularize subsequent training.

\noindent\textbf{Prototype drift.}
When adapting the backbone from $f_{t-1}$ to $f_t$ on $\mathcal{D}_t$,
the representation changes to fit the new classes and, as a side effect,
the geometry of the feature space shifts.
Hence, prototypes computed under $f_{t-1}$ become stale once $f_t$
is deployed. Denote the updated class mean, its vector displacement, and its norm by 
\vspace{-1mm}
\begin{equation}
\label{eq:vec_drift_prelim} % 
\bm{\mu}_c^{\,t}=\frac{1}{|\mathcal{D}_c|}\sum_{x\in\mathcal{D}_c} f_t(x),\quad
\bm{\Delta}_c^{\,t}=\bm{\mu}_c^{\,t}-\bm{\mu}_c^{\,t-1},\quad
\delta_c^{\,t}=\big\|\bm{\Delta}_c^{\,t}\big\|_2 .
\end{equation}
\vspace{-0.5mm}
Larger $\delta_c^{\,t}$ indicates stronger prototype drift,
which biases decisions toward recently learned classes.
Because no earlier samples are retained, $\bm{\mu}_c^{\,t}$ cannot be recomputed
exactly; mitigating or compensating for this drift under the exemplar-free constraint
motivates the two-stage strategy below. % 
\vspace{-1mm}
\subsection{Prior Drift Compensation Paradigm} \label{sec:two_stage_prior}
\vspace{-1mm}
A widely adopted recipe to handle prototype drift in EFCIL proceeds in two stages.
\vspace{-3mm}
\paragraph{Stage I: in-task regularization (backward alignment via $D$).} % 
During task $t$, the old backbone $f_{t-1}$ is frozen and used as a teacher, while the current backbone $f_t$ is trained on the new data $\mathcal{D}_t$ as the student. Let $g$ denote the classifier head (shared or task-specific). For each $x\!\in\!\mathcal{D}_t$ we define
$z_{\text{old}}=f_{t-1}(x)$, $z_{\text{new}}=f_t(x)$ and the corresponding logits
$\ell_{\text{old}}=g(z_{\text{old}})$, $\ell_{\text{new}}=g(z_{\text{new}})$.
The student is optimized with the usual cross-entropy on new labels and a
distillation/regularization term that constrains either features or logits relative to the teacher: 
\vspace{-1mm}
\begin{equation}
\mathcal{L}_{\text{S1}}
=\mathbb{E}_{(x,y)\in\mathcal{D}_t}\!
\Big[\, \mathrm{CE}(\ell_{\text{new}},y)
\;+\;\lambda\,D\big(\phi_{\text{new}}(x),\,\phi_{\text{old}}(x)\big)\Big],
\end{equation}
where $\phi$ is either $z$ (feature) or $\ell$ (logit), and $D$ stands for a distillation/regularization operator with $\lambda\!>\!0$ balancing the terms. 
This stage constrains the update of $f_t$ using only $\mathcal{D}_t$, thereby limiting the growth of $\delta_c^{\,t}$ for past classes. 
\vspace{-3mm}
\paragraph{Stage II: post-hoc prototype transport (adapter learning).}
After training $f_t$, both $f_{t-1}$ and $f_t$ are frozen and an adapter $A$ is learned on $\mathcal{D}_t$ to map old features into the new space. Concretely, $A$ is fitted on paired features $\big(f_{t-1}(x),f_t(x)\big)$ by minimizing
\vspace{-1mm}
\begin{equation}
\min_{A}\; \mathbb{E}_{x\in\mathcal{D}_t}\,\big\|A\,(f_{t-1}(x))-f_t(x)\big\|_2^2,
\end{equation}
with $A$ instantiated as a global translation operator, a class-conditioned translation, or a learnable MLP/linear projector (details vary across works; see Appendix~\ref{sec:drift_transport_view}). 
Once trained, $A$ is applied to the cached prototypes from prior steps to relocate them into the current feature space:
\vspace{-1mm}
\begin{equation}
\tilde{\bm\mu}_c^{\,t} \;=\; A\,(\bm\mu_c^{\,t-1}), \qquad c\in\mathcal{C}_{1:t-1}.
\end{equation}
These transported prototypes $\{\tilde{\bm\mu}_c^{\,t}\}$ are then used by the classifier at inference under $f_t$, effectively compensating for the shift induced by the update from $f_{t-1}$ to $f_t$. 

\textbf{Our Research Objective.}
In the two-stage paradigm, the regularization term in Stage~I (often a distillation loss) pulls the new encoder $f_t$ toward the frozen teacher $f_{t-1}$, whereas the Stage~II adapter transports old prototypes forward from the space of $f_{t-1}$ to that of $f_t$.
Functionally, these two modules act in opposite directions; structurally, a prior work~\citep{rypesc2024task} instantiates the distiller with \emph{the same architecture} as the adapter but applies it in the reverse direction ($z_{\text{new}}\!\to\! z_{\text{old}}$ vs.\ $z_{\text{old}}\!\to\! z_{\text{new}}$). 
Our objective is to make this duality explicit already in Stage~I: we co-learn a forward adapter $A$ and a backward distiller $D_t$ during Stage~I, enforcing \emph{bidirectional alignment and cycle consistency} in both function (mutual inverses on features) and structure (mirrored/tied parameters), so that prototype transport becomes more accurate by design.

% \begin{figure}[t]
% \centering
% \begin{minipage}{0.99\columnwidth}
%   \captionsetup{width=\linewidth,justification=centering,singlelinecheck=true}
%   \centering
%   \includegraphics[width=\linewidth]{figs/main_figure.pdf}
%     \caption{\textbf{Overview.} \textbf{(1) Train:} the current backbone $f_t$
%     learns on $\mathcal{D}_t$ (producing $z_{\text{new}}$, while frozen $f_{t-1}$
%     provides $z_{\text{old}}$) with task loss $\mathcal{L}_{\mathrm{CE}}$.
%     \textbf{(2) Bidirectional alignment:} jointly learn a distiller
%     $D\!:z_{\text{new}}\!\to\!z_{\text{old}}$ and an adapter
%     $A\!:z_{\text{old}}\!\to\!z_{\text{new}}$ using $\mathcal{L}_{\mathrm{bi}}$.
%     \textbf{(3) Cycle consistency:} enforce $A\!\circ\!D\!\approx\!I$ and
%     $D\!\circ\!A\!\approx\!I$ with $\mathcal{L}_{\mathrm{cyc}}$ (targets detached),
%     yielding near-bijective, geometry-preserving transport. Old Gaussian prototypes
%     are mapped forward by $A$, and all classes are evaluated in the \emph{new} space.}

%   \label{fig:loss_ablation_right}
% \end{minipage}
% \end{figure}

\vspace{-2mm}
\section{Methodology}
\vspace{-2mm}
\label{Method}

\subsection{Setup}

Let $f_{t-1}$ be the frozen old backbone from task $t{-}1$ and $f_t$ the backbone being trained at task $t$. For an input $x$,
\[
z_{\text{old}}=f_{t-1}(x)\in\mathbb{R}^d,\qquad
z_{\text{new}}=f_t(x)\in\mathbb{R}^d.
\]
Unless otherwise noted, evaluation is performed \emph{in the new feature space} of $f_t$ using a Bayes classifier (see Appendix~A.1.2), with class statistics estimated from $\mathcal{D}_t$ (new classes) or transported into the new space (old classes). We instantiate two lightweight maps: a \textbf{distiller} $D:\mathbb{R}^d\!\to\!\mathbb{R}^d$ (new$\!\to$old) and an \textbf{adapter} $A:\mathbb{R}^d\!\to\!\mathbb{R}^d$ (old$\!\to$new), implemented as linear layers or shallow MLPs. We use the stop-gradient operator $\stopgrad(\cdot)$ throughout.

\subsection{Joint Training with Bidirectional Cycle Consistency}
\label{sec:Joint_Training_with_bicyc}

We train $f_t$, $A$, and $D$ jointly on $\mathcal{D}_t$, combining standard classification with teacher–student regularization and our bidirectional/cycle consistency. Let $g$ be the task-specific classifier head with logits $\ell_{\text{new}}=g(z_{\text{new}})$. For brevity, we denote the bidirectional alignment + consistency cycle module as \textbf{Bicyc}$(z_{\text{old}}, z_{\text{new}})$ (see Fig.~\ref{fig:main_pipline}).

\vspace{-2mm}
\paragraph{Bidirectional alignment.}
We seek (i) \emph{backward compatibility} by making $z_{\text{new}}$ projectable to the old space via $D$, and (ii) a \emph{forward} map $A$ that transports old prototypes into the current space used for evaluation—without dragging $f_t$ backward. Concretely,
\vspace{-1mm}
\begin{equation}
\label{eq:lbi}
\mathcal{L}_{\text{bi}}
=
\big\|D(z_{\text{new}})-z_{\text{old}}\big\|_2^2
+
\big\|A(z_{\text{old}})-\stopgrad(z_{\text{new}})\big\|_2^2.
\end{equation}
The first term updates $f_t$ and $D$ (feature-level distillation, new$\!\to$old). The second term updates $A$ only (detached target), so $A$ \emph{chases} the evolving new space (old$\!\to$new) without reducing the plasticity of $f_t$. In a linear–Gaussian view, minimizing \eqref{eq:lbi} reduces transport errors $\varepsilon_{\text{old}\to\text{new}}=\mathbb{E}\|A(z_{\text{old}})-z_{\text{new}}\|^2$ and $\varepsilon_{\text{new}\to\text{old}}=\mathbb{E}\|D(z_{\text{new}})-z_{\text{old}}\|^2$, which bound prototype mean/covariance mismatch after transport and help control margin drift.

\vspace{-2mm}
\paragraph{Cycle consistency.}
While $\mathcal{L}_{\text{bi}}$ aligns both directions, it does not by itself prevent degeneracies (e.g., rank loss in weakly correlated directions). We therefore add a cycle loss that nudges the compositions toward identity on the data support:
\vspace{-1mm}
\begin{equation}
\label{eq:lcyc}
\mathcal{L}_{\text{cyc}}
=
\big\|A\!\big(D(z_{\text{new}})\big)-\stopgrad(z_{\text{new}})\big\|_2^2
+
\big\|D\!\big(A(z_{\text{old}})\big)-\stopgrad(z_{\text{old}})\big\|_2^2.
\end{equation}

Targets are detached, so $\mathcal{L}_{\text{cyc}}$ \emph{stabilizes} $(A,D)$ without pulling $f_t$. Spectrally, enforcing $A\!\circ\!D\!\approx\!I$ and $D\!\circ\!A\!\approx\!I$ contracts the singular values of the composed transports toward $1$, curbing rank/energy loss and promoting near-isometric geometry preservation. Thus $\mathcal{L}_{\text{bi}}$ lowers transport error (alignment) while $\mathcal{L}_{\text{cyc}}$ regularizes the transport \emph{operators} (near-bijection); together they yield faithful prototype transport and empirically reduce forgetting without sacrificing plasticity. We denote the weighted sum of the bidirectional alignment and cycle-consistency losses by:
\begin{equation}
\text{\textbf{Bicyc}}(z_{\text{old}}, z_{\text{new}})
\coloneqq
\lambda_{\text{bi}}\,\mathcal{L}_{\text{bi}}
+ \lambda_{\text{cyc}}\,\mathcal{L}_{\text{cyc}}
\end{equation}
\noindent We analyze the cycle objective under centered features and full-rank covariances on the data support, passing to whitened variables $\tilde z_{\text{old}}=\Sigma_{\text{old}}^{-1/2}z_{\text{old}}$ and $\tilde z_{\text{new}}=\Sigma_{\text{new}}^{-1/2}z_{\text{new}}$. In this space $\EE[\tilde z_{\text{new}}\tilde z_{\text{new}}^\top]=I$, and the expected cycle error equals the squared Frobenius distance of $\tilde A\tilde D$ to $I$. We now state the resulting contraction property.

\textbf{Theorem 1 (Cycle contraction).} Let $\Sigma_{\text{old}}=\EE[z_{\text{old}}z_{\text{old}}^\top]$ and $\Sigma_{\text{new}}=\EE[z_{\text{new}}z_{\text{new}}^\top]$ be full-rank on the data support and define whitened variables $\tilde z_{\text{old}}=\Sigma_{\text{old}}^{-1/2}z_{\text{old}}$, $\tilde z_{\text{new}}=\Sigma_{\text{new}}^{-1/2}z_{\text{new}}$ with induced maps $\tilde A=\Sigma_{\text{new}}^{-1/2}A\Sigma_{\text{old}}^{1/2}$ and $\tilde D=\Sigma_{\text{old}}^{-1/2}D\Sigma_{\text{new}}^{1/2}$. Let $M := \tilde A \tilde D - I$. If the features are centered, then:
\begin{equation}
\mathbb{E}\,\bigl\| M \tilde z_{\mathrm{new}} \bigr\|_2^2 = \| M \|_F^2.
\end{equation}

By Mirsky/Hoffman--Wielandt \citep{HornJohnson2013} $\sum_{k=1}^{d} \bigl(\sigma_k(\tilde A \tilde D) - 1\bigr)^2 \;\le\; \|M\|_F^2$ and hence $\max_{k}\, \bigl| \sigma_k(\tilde A \tilde D) - 1 \bigr| \;\le\; \|M\|_F$. In particular, if $\|M\|_2 < 1$ then $1 - \|M\|_2 \;\le\; \sigma_k(\tilde A \tilde D) \;\le\; 1 + \|M\|_2$ and $\kappa(\tilde A \tilde D) \;\le\; \frac{1 + \|M\|_2}{1 - \|M\|_2}$. Consequently, minimizing $\mathcal{L}_{\text{cyc}}$ drives the singular values of $\tilde A\tilde D$ toward $1$ on the data support, preventing rank loss and preserving geometry. Proof in Appendix~\ref{sec:proof_theorem1}.

% Then
% \begin{equation}
% \EE\!\left\|(\tilde A\tilde D-I)\,\tilde z_{\text{new}}\right\|_2^2
% =\|\tilde A\tilde D-I\|_F^2.
% \end{equation}

\textbf{Corollary 2 (Decision stability for classification).} Let old-class statistics be transported as $\hat\mu_c^{\,t}=A\mu_c^{\,t-1}$ and (for linear $A$) $\hat\Sigma_c^{\,t}=A\Sigma_c^{\,t-1}A^\top$.
Assume evaluation uses the Bayes rule with Gaussian class-conditionals $(\mu_c^t,\Sigma_c^t)$ and priors $\pi_c$, with log-scores $\ell_c(x)$ as in Appendix~\ref{sec:bayes_classifier}.
Define mean transport errors $\delta_c:=\|\hat\mu_c^{\,t}-\mu_c^t\|_{(\Sigma_c^t)^{-1}}$.
If the alignment error $\varepsilon_{\text{old}\to\text{new}}^2=\EE\|A z_{\text{old}}-z_{\text{new}}\|_2^2$ and the cycle error $\varepsilon_{\text{cyc,new}}^2=\EE\|AD z_{\text{new}}-z_{\text{new}}\|_2^2$ are small, then:
\begin{equation}
\delta_c \;\lesssim\; \sqrt{\varepsilon_{\text{old}\to\text{new}}^2}\,,
\qquad
\|\tilde\Sigma^{\,t}-\Sigma^t\|_2 \;\lesssim\; C_1\sqrt{\varepsilon_{\text{old}\to\text{new}}^2}+C_2\,\varepsilon_{\text{cyc,new}}.
\end{equation}
For any class pair $(i,j)$ and any $x$, let $m_{ij}(x):=|\ell_i(x)-\ell_j(x)|$ be the Bayes margin.
Then the induced change in log-odds satisfies
$\big|(\hat\ell_i-\hat\ell_j)-(\ell_i-\ell_j)\big|\;\lesssim\; C_\mu(\delta_i+\delta_j)+C_\Sigma\|\hat\Sigma^{\,t}-\Sigma^t\|_2$.
Consequently, if $C_\mu(\delta_i+\delta_j)+C_\Sigma\|\hat\Sigma^{\,t}-\Sigma^t\|_2 < m_{ij}(x)$, the Bayes decision between $i$ and $j$ at $x$ remains unchanged after transport. Proof in Appendix~\ref{sec:proof_corollary2}.

\vspace{-2mm}
\paragraph{Pitfall of anti-collapse loss.}
For features $z\in\mathbb{R}^{B\times S}$, let $\Sigma=\frac{1}{B-1}(z-\bar z)^\top(z-\bar z)$. The AdaGauss anti-collapse loss~\citep{rypesc2024task} is
\begin{equation}
\mathcal{L}_{\mathrm{ac}}=-\tfrac{1}{S}\sum_{i=1}^S \min\!\big(\mathrm{chol}(\Sigma)_{ii},\,\beta\big).
\end{equation}
In practice, mini-batch $\Sigma$ can be non-SPD or rank-deficient, causing Cholesky failures and potentially inflating scale near ill-conditioning. We enforce SPD via symmetrization and shrinkage, with a jittered Cholesky and eigenvalue flooring as fallback:
\begin{equation}
\tilde{\Sigma}=\tfrac{1}{2}\big(\Sigma+\Sigma^\top\big),\qquad
\hat{\Sigma}=\tilde{\Sigma}+\lambda\,\tfrac{\mathrm{tr}(\tilde{\Sigma})}{S}\,I+\varepsilon I,
\end{equation}
and, for very small batches, we optionally use a diagonal approximation $\hat{\Sigma}_{\mathrm{diag}}=\operatorname{diag}(\operatorname{diag}(\hat{\Sigma}))$. The robust objective is
\begin{equation}
\mathcal{L}_{\mathrm{ac}}^{\mathrm{rob}}=-\tfrac{1}{S}\sum_{i=1}^S \min\!\big(\mathrm{chol}(\hat{\Sigma})_{ii},\,\beta\big).
\end{equation}

% \vspace{-2mm}
% \paragraph{Pitfall of anti-collapse loss.}
% For features $z\in\mathbb{R}^{B\times S}$, let $\Sigma=\frac{1}{B-1}(z-\bar z)^\top(z-\bar z)$.
% We use $\mathrm{chol}(\Sigma)$ to denote the lower–triangular Cholesky factor $L$ such that $\Sigma=LL^\top$,
% and write $(\mathrm{chol}(\Sigma))_{ii}:=L_{ii}$ for its $i$-th diagonal entry ($i=1,\dots,S$).
% The AdaGauss anti-collapse loss~\citep{rypesc2024task} is
% \begin{equation}
% \mathcal{L}_{\mathrm{ac}}=-\tfrac{1}{S}\sum_{i=1}^S \min\!\big((\mathrm{chol}(\Sigma))_{ii},\,\beta\big).
% \end{equation}
% In practice, mini-batch $\Sigma$ can be non-SPD or rank-deficient, causing Cholesky failures and potentially inflating scale near ill-conditioning. We enforce SPD via symmetrization and shrinkage, with a jittered Cholesky and eigenvalue flooring as fallback:
% \begin{equation}
% \tilde{\Sigma}=\tfrac{1}{2}\big(\Sigma+\Sigma^\top\big),\qquad
% \hat{\Sigma}=\tilde{\Sigma}+\lambda\,\tfrac{\mathrm{tr}(\tilde{\Sigma})}{S}\,I+\varepsilon I,
% \end{equation}
% and, for very small batches, we optionally use a diagonal approximation $\hat{\Sigma}_{\mathrm{diag}}=\operatorname{diag}(\operatorname{diag}(\hat{\Sigma}))$. The robust objective is
% \begin{equation}
% \mathcal{L}_{\mathrm{ac}}^{\mathrm{rob}}=-\tfrac{1}{S}\sum_{i=1}^S \min\!\big((\mathrm{chol}(\hat{\Sigma}))_{ii},\,\beta\big).
% \end{equation}

\vspace{-2mm}
\paragraph{Total Stage-I loss and gradient routing.} Combining the classification, cycle, and anti-collapse terms yields:
\vspace{-1mm}
\begin{equation}
\label{eq:ltotal}
\mathcal{L}_{\text{total}}
=
\underbrace{\mathcal{L}_{\text{CE}}(\ell_{\text{new}},y)}_{\text{learn new classes}}
+ \text{\textbf{Bicyc}}(z_{\text{old}}, z_{\text{new}})
+\alpha\,\mathcal{L}_{\mathrm{ac}}^{\mathrm{rob}}.
\end{equation}
Here, $\mathcal{L}_{\text{CE}}$ and the first term of \eqref{eq:lbi} update $f_t$ (and $D$); the second term of \eqref{eq:lbi} updates $A$ only (detached target); and \eqref{eq:lcyc} stabilizes $(A,D)$ without reducing the plasticity of $f_t$. Importantly, if gradients from the adapter are allowed to flow into $f_t$, $A$ and $D$ become adversarial, severely weakening $D$’s regularization and causing sharp performance drops. After Stage~I, we freeze $f_{t-1}$, $f_t$, and $D$, and perform a low-learning-rate fine-tuning of $A$ on $\mathcal{D}_t$ to sharpen transport without re-optimizing from scratch.

\begin{table*}[t]
\caption{Average incremental ($A_{\text{inc}}$) and last-task average ($A_{\text{last}}$) accuracy (\%, mean $\pm$ std.\ over five runs) on CIFAR-100 and TinyImageNet when training the feature extractor from scratch. Best results are \textbf{bold}.}
\label{tab:main_results}
\centering
% Optional: tighten columns a bit for ICLR two-column layout
\setlength{\tabcolsep}{3pt}
\small
\begin{tabular}{lcccccccc}
\toprule
\multirow{3}{*}{\textbf{Method}} &
  \multicolumn{4}{c}{\textbf{CIFAR-100}} &
  \multicolumn{4}{c}{\textbf{TinyImageNet}} \\
\cmidrule(lr){2-5}\cmidrule(lr){6-9}
  & \multicolumn{2}{c}{$T{=}10$} &
    \multicolumn{2}{c}{$T{=}20$} &
    \multicolumn{2}{c}{$T{=}10$} &
    \multicolumn{2}{c}{$T{=}20$} \\
\cmidrule(lr){2-3}\cmidrule(lr){4-5}
\cmidrule(lr){6-7}\cmidrule(lr){8-9}
  & $A_{\text{last}}$ & $A_{\text{inc}}$
  & $A_{\text{last}}$ & $A_{\text{inc}}$
  & $A_{\text{last}}$ & $A_{\text{inc}}$
  & $A_{\text{last}}$ & $A_{\text{inc}}$ \\
\midrule
EWC &
  30.9$\pm$1.9 & 50.4$\pm$1.7 &
  17.0$\pm$1.6 & 34.2$\pm$2.1 &
  18.5$\pm$1.8 & 34.3$\pm$2.3 &
  11.3$\pm$1.9 & 26.8$\pm$2.5 \\[2pt]

LwF\conf{ECCV16} &
  31.9$\pm$1.1 & 51.8$\pm$1.5 &
  17.6$\pm$1.2 & 39.2$\pm$1.7 &
  27.1$\pm$1.5 & 39.6$\pm$2.0 &
  15.2$\pm$1.6 & 31.5$\pm$2.1 \\[2pt]

SDC\conf{CVPR20} &
  40.6$\pm$0.9 & 56.2$\pm$1.3 &
  32.3$\pm$1.0 & 46.6$\pm$1.4 &
  29.5$\pm$1.1 & 43.8$\pm$1.5 &
  26.3$\pm$1.2 & 40.6$\pm$1.7 \\[2pt]

PASS\conf{CVPR21} &
  30.8$\pm$1.2 & 48.3$\pm$1.1 &
  17.6$\pm$0.8 & 31.1$\pm$1.3 &
  24.5$\pm$0.6 & 39.5$\pm$1.0 &
  18.5$\pm$1.4 & 30.4$\pm$1.9 \\[2pt]

FeTrIL\conf{WACV23} &
  34.9$\pm$0.5 & 51.2$\pm$1.1 &
  23.3$\pm$1.4 & 37.9$\pm$1.2 &
  31.0$\pm$0.9 & 45.3$\pm$1.8 &
  25.9$\pm$1.2 & 39.9$\pm$1.2 \\[2pt]

FeCAM\conf{NeurIPS23} &
  32.4$\pm$0.5 & 48.7$\pm$0.9 &
  21.1$\pm$1.0 & 34.5$\pm$1.3 &
  30.9$\pm$0.9 & 44.9$\pm$1.4 &
  24.9$\pm$0.8 & 37.9$\pm$1.4 \\[2pt]

EFC\conf{ICLR24} &
  43.5$\pm$0.8 & 58.1$\pm$1.2 &
  32.4$\pm$0.9 & 47.0$\pm$1.3 &
  34.5$\pm$1.1 & 47.9$\pm$1.5 &
  28.4$\pm$1.2 & 42.1$\pm$1.6 \\[2pt]

ADC\conf{CVPR24} &
  46.5$\pm$1.2 & 61.4$\pm$1.6 &
  35.1$\pm$1.4 & 51.7$\pm$1.8 &
  32.3$\pm$1.5 & 43.0$\pm$1.9 &
  18.1$\pm$1.6 & 36.0$\pm$2.1 \\[2pt]

LDC\conf{ECCV24} &
  45.4$\pm$1.6 & 59.5$\pm$1.9 &
  35.5$\pm$1.9 & 51.9$\pm$2.3 &
  34.2$\pm$1.1 & 46.8$\pm$1.6 &
  24.9$\pm$2.2 & 38.2$\pm$2.7 \\[2pt]

AdaGauss\conf{NeurIPS24} &
  46.8$\pm$1.2 & 60.9$\pm$1.0 &
  37.9$\pm$1.0 & 54.4$\pm$0.8 &
  32.9$\pm$0.9 & 45.8$\pm$1.3 &
  27.5$\pm$1.2 & 39.5$\pm$1.1 \\[2pt]

% DPCR\conf{NeurIPS24} &
%   46.8 & 60.9 &
%   37.9& 54.4&
%   32.9& 45.8&
%   27.5& 39.5 \\[2pt]

DPCR\conf{ICML2025} &
50.2$\pm$0.7 & 62.8$\pm$1.1 &
39.8$\pm$1.2 & 54.8$\pm$0.9 &
34.3$\pm$1.8 & 46.9$\pm$0.9 &
25.6$\pm$0.7 & 39.3$\pm$0.6 \\[2pt]

\textbf{BiCyc (Ours)} &
  \textbf{50.6$\pm$0.9} & \textbf{63.2$\pm$1.3} &
  \textbf{41.5$\pm$1.1} & \textbf{56.5$\pm$1.3} &
  \textbf{35.4$\pm$0.8} & \textbf{49.1$\pm$1.4} &
  \textbf{30.2$\pm$1.1} & \textbf{44.2$\pm$1.3} \\
\bottomrule
\end{tabular}
\end{table*}

\begin{table*}[t]
\caption{Average incremental ($A_{\text{inc}}$) and last-task average ($A_{\text{last}}$) accuracy (\%, mean $\pm$ std.\ over five runs) on ImageNet-100 and CUB-200. Best results are \textbf{bold}. \textsuperscript{\dag}: results excerpted from~\citep{GomezVillaGWBTW24}. \textsuperscript{\ddag}: results excerpted from~\citep{hesemantic}.}
\label{tab:main_results2}
\centering
% Optional: tighten columns if needed
\setlength{\tabcolsep}{2.1pt}
\small
\begin{tabular}{lcccccccc}
\toprule
\multirow{3}{*}{\textbf{Method}} &
  \multicolumn{4}{c}{\textbf{ImageNet-100}} &
  \multicolumn{4}{c}{\textbf{CUB-200}} \\
\cmidrule(lr){2-5}\cmidrule(lr){6-9}
  & \multicolumn{2}{c}{$T{=}10$} &
    \multicolumn{2}{c}{$T{=}20$} &
    \multicolumn{2}{c}{$T{=}10$} &
    \multicolumn{2}{c}{$T{=}20$} \\
\cmidrule(lr){2-3}\cmidrule(lr){4-5}
\cmidrule(lr){6-7}\cmidrule(lr){8-9}
  & $A_{\text{last}}$ & $A_{\text{inc}}$
  & $A_{\text{last}}$ & $A_{\text{inc}}$
  & $A_{\text{last}}$ & $A_{\text{inc}}$
  & $A_{\text{last}}$ & $A_{\text{inc}}$ \\
\midrule
EWC &
  25.1$\pm$2.8 & 40.6$\pm$3.3 &
  13.7$\pm$2.1 & 29.2$\pm$2.5 &
  15.8$\pm$0.7 & 32.6$\pm$0.5 &
  12.3$\pm$0.8 & 27.2$\pm$0.6 \\[2pt]

LwF\conf{ECCV16} &
  33.4$\pm$2.2 & 51.5$\pm$1.6 &
  18.6$\pm$1.6 & 41.3$\pm$1.9 &
  30.4$\pm$1.1 & 46.1$\pm$1.0 &
  19.4$\pm$1.6 & 34.7$\pm$1.8 \\[2pt]

SDC\conf{CVPR20} &
  35.4$\pm$1.9 & 50.1$\pm$1.6 &
  19.4$\pm$1.0 & 36.5$\pm$1.4 &
  50.3$\pm$1.3 & 60.5$\pm$1.2 &
  27.9$\pm$1.4 & 40.1$\pm$1.6 \\[2pt]

PASS\conf{CVPR21} &
  26.4$\pm$1.3 & 45.7$\pm$0.2 &
  14.4$\pm$1.2 & 31.7$\pm$0.4 &
  27.0$\pm$0.9 & 42.3$\pm$0.9 &
  18.1$\pm$1.2 & 36.9$\pm$1.1 \\[2pt]

FeTrIL\conf{WACV23} &
  36.2$\pm$1.2 & 52.6$\pm$0.6 &
  26.6$\pm$1.5 & 42.4$\pm$2.1 &
  36.9$\pm$0.7 & 48.2$\pm$0.6 &
  34.6$\pm$1.0 & 45.3$\pm$0.9 \\[2pt]

FeCAM\conf{NeurIPS23} &
  38.7$\pm$1.0 & 54.8$\pm$0.5 &
  29.0$\pm$1.3 & 44.6$\pm$2.0 &
  40.2$\pm$0.8 & 54.9$\pm$1.0 &
  36.2$\pm$1.1 & 48.9$\pm$1.3 \\[2pt]

EFC\conf{ICLR24} &
  50.9$\pm$1.1 & 61.3$\pm$1.2 &
  38.6$\pm$1.2 & 50.5$\pm$1.5 &
  51.0$\pm$0.6 & 63.3$\pm$0.7 &
  \textbf{46.1$\pm$1.0} & \textbf{59.3$\pm$1.3} \\[2pt]

ADC\conf{CVPR24} &
  38.3$\pm$1.2 & 55.5$\pm$1.5 &
  25.1$\pm$1.3 & 43.4$\pm$1.7 &
  49.5$\pm$0.9 & 58.8$\pm$1.1 &
  35.4$\pm$1.4 & 48.3$\pm$1.4 \\[2pt]

LDC\conf{ECCV24} &
  51.4\textsuperscript{\dag}$\pm$1.2\textsuperscript{\dag} & \textbf{69.4\textsuperscript{\dag}$\pm$0.6\textsuperscript{\dag}} &
  28.5$\pm$1.7 & 46.5$\pm$2.7 &
  47.5$\pm$0.7 & 55.7$\pm$1.3 &
  27.2$\pm$1.1 & 39.8$\pm$2.1 \\[2pt]

AdaGauss\conf{NeurIPS24} &
  51.1$\pm$1.2 & 65.0$\pm$1.4 &
  42.6$\pm$1.6 & 57.4$\pm$1.9 &
  52.9$\pm$0.8 & 63.4$\pm$1.3 &
 45.0$\pm$1.3 & 57.0$\pm$1.0 \\[2pt]

DPCR\conf{ICML2025} &
49.9$\pm$0.8 & 64.8$\pm$1.1 &
37.3$\pm$1.6 & 54.7$\pm$0.7 &
-- & -- &
-- & -- \\[2pt]

\textbf{BiCyc (Ours)} &
  \textbf{52.7$\pm$0.9} & 66.8$\pm$1.4 &
  \textbf{43.8$\pm$1.4} & \textbf{58.2$\pm$1.8} &
  \textbf{53.7$\pm$0.7} & \textbf{64.0$\pm$0.8} &
  43.7$\pm$1.4 & 55.9$\pm$1.2 \\
\bottomrule
\end{tabular}
\vspace{-2mm}
\end{table*}

\vspace{-2mm}
\section{Experiments}
\vspace{-2mm}

\begin{table*}[t]
\caption{Last-task average forgetting ($F_{\text{last}}$) (\%, mean $\pm$ std.\ over five runs). CIFAR-100, TinyImageNet, and ImageNet-100 use from-scratch training; CUB-200 uses ImageNet-pretrained initialization. Best results are \textbf{bold}.}
\label{tab:main_results3}
\centering
\setlength{\tabcolsep}{2.8pt}
\small
\providecommand{\NA}{\textemdash}
\begin{tabular}{lcccccccc}
\toprule
\multirow{3}{*}{\textbf{Method}} &
  \multicolumn{2}{c}{\textbf{CIFAR-100}} &
  \multicolumn{2}{c}{\textbf{TinyImageNet}} &
  \multicolumn{2}{c}{\textbf{ImageNet-100}} &
  \multicolumn{2}{c}{\textbf{CUB-200}} \\
\cmidrule(lr){2-3}\cmidrule(lr){4-5}\cmidrule(lr){6-7}\cmidrule(lr){8-9}
  & $T{=}10$ & $T{=}20$ & $T{=}10$ & $T{=}20$ & $T{=}10$ & $T{=}20$ & $T{=}10$ & $T{=}20$ \\
\cmidrule(l{0.5em}r{0.5em}){2-3}\cmidrule(l{0.5em}r{0.5em}){4-5}\cmidrule(l{0.5em}r{0.5em}){6-7}\cmidrule(l{0.5em}r{0.5em}){8-9}
  & $F_{\text{last}}$ & $F_{\text{last}}$ & $F_{\text{last}}$ & $F_{\text{last}}$ & $F_{\text{last}}$ & $F_{\text{last}}$ & $F_{\text{last}}$ & $F_{\text{last}}$ \\
\midrule
  
LwF\conf{ECCV16} &
  23.2$\pm$1.7 & 31.2$\pm$1.8 &
  21.9$\pm$1.9 & 33.5$\pm$2.4 &
  42.1$\pm$2.3 & 48.1$\pm$2.2 &
  16.5$\pm$1.1 & 21.7$\pm$1.4 \\[2pt]

SDC\conf{CVPR20} &
  34.8$\pm$1.7 & 35.9$\pm$1.9 &
  25.1$\pm$1.4 & 29.4$\pm$2.1 &
  44.6$\pm$2.0 & 54.4$\pm$2.3 &
  10.9$\pm$1.3 & 17.3$\pm$1.1 \\[2pt]

EFC\conf{ICLR24} &
  23.1$\pm$1.1 & 24.7$\pm$1.8 &
  23.5$\pm$2.4 & 30.1$\pm$3.0 &
  21.5$\pm$1.9 & 23.8$\pm$2.5 &
  \textbf{10.7$\pm$0.7} & \textbf{14.8$\pm$1.7} \\[2pt]

ADC\conf{CVPR24} &
  21.9$\pm$1.1 & 31.0$\pm$1.6 &
  30.2$\pm$2.0 & 36.8$\pm$1.9 &
  32.4$\pm$1.6 & 33.4$\pm$1.8 &
  12.8$\pm$1.1 & 21.3$\pm$1.5 \\[2pt]

LDC\conf{ECCV24} &
  21.7$\pm$1.9 & 25.6$\pm$2.3 &
  24.7$\pm$2.5 & 30.7$\pm$2.1 &
  25.7$\pm$1.7 & 32.9$\pm$2.3 &
  13.6$\pm$1.2 & 23.9$\pm$1.8 \\[2pt]

AdaGauss\conf{NeurIPS24} &
  16.7$\pm$1.4 & 21.0$\pm$1.5 &
  18.7$\pm$1.2 & 23.1$\pm$1.0 &
  20.6$\pm$0.9 & 22.9$\pm$1.1 & 
  11.6$\pm$0.7 & 16.9$\pm$1.3 \\[2pt]

\textbf{BiCyc (Ours)} &
  \textbf{13.5$\pm$1.3} & \textbf{16.6$\pm$0.9} &
  \textbf{12.0$\pm$0.9} & \textbf{18.9$\pm$1.1} &
  \textbf{18.2$\pm$1.6} & \textbf{20.8$\pm$1.4} & 
  11.3$\pm$0.9 & 17.5$\pm$1.3 \\
\bottomrule
\end{tabular}
\vspace{-8mm}
\end{table*}

% \section{Drift Metrics Between Stored Prototypes and Test-Time Estimates}

\textbf{Baselines.}
We benchmark our approach against a broad set of exemplar-free class-incremental learning (EFCIL) methods.%
\ Classic regularization baselines—EWC~\citep{kirkpatrick2017ewc} and LwF~\citep{LiH16}—are executed using the reference \textsc{OCL} implementation~\citep{MAI202228}.%
\ Contemporary state-of-the-art approaches—SDC~\citep{0004TLHWCJ020}, PASS~\citep{zhu2021pass}, FeTrIL~\citep{petit2023fetril}, FeCAM~\citep{toldo2022fecam}, EFC~\citep{Magistri2024EFC}, ADC~\citep{GoswamiSLKT024}, LDC~\citep{GomezVillaGWBTW24}, and AdaGauss~\citep{rypesc2024task}—are run with the authors’ public codebases as distributed via FACIL~\citep{masana2022class}, PyCIL~\citep{zhou2023pycil}, or the official repositories.%
\ Unless otherwise noted, we preserve the original data augmentations and default hyper-parameters reported by each paper.

\textbf{Implementation details and reproducibility.}
We build on the public AdaGauss codebase and add the components introduced in this work.
Unless stated otherwise, all experiments use a ResNet\mbox{-}18 backbone with a batch size of 256 images per iteration, following AdaGauss.
For CIFAR\mbox{-}100~\citep{krizhevsky2009learning}, TinyImageNet~\citep{le2015tiny}, and ImageNet\mbox{-}100~\citep{5206848}, the backbone is trained from scratch for 200 epochs using SGD with a fixed initial learning rate of \(1\times10^{-1}\) and weight decay \(5\times10^{-4}\); the learning rate is decayed by a factor of 10 at epochs \(\{60,120,180\}\).
For CUB\mbox{-}200~\citep{wah2011caltech}, we initialize from ImageNet-pretrained weights and adopt a split learning rate: \(1\times10^{-2}\) for the backbone and \(1\times10^{-1}\) for the heads.
The distiller and adapter are trained with learning rate \(5\times10^{-2}\) and weight decay \(1\times10^{-4}\).
For the main from-scratch experiments, we set \(\lambda_{\text{bi}}{=}5\) and \(\lambda_{\text{cyc}}{=}1\).
After Stage~I, we fine\mbox{-}tune the adapter for 30 epochs using SGD with an initial learning rate of \(1\times10^{-2}\) and weight decay \(5\times10^{-4}\).

All other hyperparameters follow AdaGauss verbatim. In particular, we keep its prototype storage and sampling settings unchanged; the small additional adapter overhead is quantified in Sec.~\ref{sec:overhead_ada64}.
For completeness, we note that the public AdaGauss code reports TinyImageNet results averaged over splits formed from the \emph{first} 100 classes, which is slightly misaligned with common balanced partitions.
To enable an apples\mbox{-}to\mbox{-}apples comparison, our tables present the corrected numbers under the standard balanced partitioning.

\textbf{Evaluation metrics.}
We report three standard measures: the last-task average accuracy \(A_{\text{last}}\), the average incremental accuracy \(A_{\text{inc}}\), defined as the running mean of \(A_{\text{last}}\) over incremental steps, and the last-task average forgetting \(F_{\text{last}}\). Dataset specifics, hyper-parameter schedules, and metric definitions are provided in the Appendix~\ref{sec:acc_metrics}.

\vspace{-1mm}
\subsection{Main Results}
\label{sec:main_results}
\vspace{-1mm}
Tables~\ref{tab:main_results} and \ref{tab:main_results2} report results on CIFAR-100, TinyImageNet, and ImageNet-100 with feature extractors trained from scratch, and on CUB-200 with ImageNet-pretrained initialization (mean$\pm$std over five runs).
\textbf{CIFAR--100:} compared to AdaGauss, we gain \textbf{+3.8}/\textbf{+2.3} pp at $T{=}10$ and \textbf{+3.6}/\textbf{+2.1} pp at $T{=}20$. 
DPCR\textsuperscript{\ddag} is competitive, but our method still slightly leads on all CIFAR-100 settings (e.g., +0.4/+0.4 pp at $T{=}10$ and +1.7/+1.7 pp at $T{=}20$).
\textbf{TinyImageNet:} improvements over AdaGauss are \textbf{+2.5}/\textbf{+3.3} pp at $T{=}10$ and \textbf{+2.7}/\textbf{+4.7} pp at $T{=}20$; the margins over the second-best (EFC) are +0.9/+1.2 pp ($T{=}10$) and +1.8/+2.1 pp ($T{=}20$). 
DPCR again trails our method, with gaps of about +1.1/+2.2 pp at $T{=}10$ and +4.6/+4.9 pp at $T{=}20$.
\textbf{ImageNet--100:} vs.\ AdaGauss we obtain \textbf{+1.6}/\textbf{+1.8} pp at $T{=}10$ and \textbf{+1.2}/\textbf{+0.8} pp at $T{=}20$; at $T{=}10$ our $A_{\text{last}}$ is best (runner-up LDC$^{\dagger}$, +1.3 pp), while $A_{\text{inc}}$ is 2.6 pp below the best (LDC$^{\dagger}$). \ul{Under our protocol, rerunning public LDC code at $T{=}10$ yields $A_{\text{last}}{=}41.7\pm1.5\%$ and $A_{\text{inc}}{=}58.7\pm1.7\%$.} DPCR\textsuperscript{\ddag} is clearly weaker than AdaGauss and ours: at $T{=}10$ it trails our method by about 2.8/2.0 pp in $A_{\text{last}}/A_{\text{inc}}$ (and is already slightly below AdaGauss by 1.2/0.2 pp), while at $T{=}20$ the gap to ours further widens to 6.5/3.5 pp (with AdaGauss still ahead of DPCR by 5.3/2.7 pp).
For $T{=}20$ we achieve the best $A_{\text{last}}$ and $A_{\text{inc}}$ (runner-up AdaGauss: +1.2/+0.8 pp). 
\textbf{CUB--200 (ImageNet pre-trained):} our performance is close to AdaGauss (vs.\ AdaGauss: +0.8/+0.6 pp at $T{=}10$, $-1.3$/$-1.1$ pp at $T{=}20$), while on the 20-split setting we trail EFC by 2.4/3.4 pp. 
DPCR does not report its CUB-200 hyperparameter configuration under our training protocol, so the corresponding entries are marked ``--'' in Table~\ref{tab:main_results2}.
With a pretrained backbone, practitioners typically adopt a very low backbone learning rate, which keeps cross-task feature drift small and thus limits the incremental gains of our method.

\begin{wraptable}[9]{r}{0.48\columnwidth}
  \vspace{-0.4cm} 
  \caption{CIFAR-100: Contributions of $\mathcal{L}_{\text{bi}}$ and $\mathcal{L}_{\text{cyc}}$.}
  \label{tab:model_component}
  \centering
  \small
  \setlength{\tabcolsep}{4pt}
  \renewcommand{\arraystretch}{0.95}
  \resizebox{\linewidth}{!}{%
    \begin{tabular}{@{}cc|cc|cc@{}}
      \toprule
      \multicolumn{2}{c|}{\textbf{Components}} &
        \multicolumn{2}{c|}{$T{=}10$} &
        \multicolumn{2}{c}{$T{=}20$} \\
      \cmidrule(lr){1-2}\cmidrule(lr){3-4}\cmidrule(lr){5-6}
      $\mathcal{L}_{\text{bi}}$ & $\mathcal{L}_{\text{cyc}}$ &
      \(A_{\text{last}}\)(\%) & \(A_{\text{inc}}\)(\%) &
      \(A_{\text{last}}\)(\%) & \(A_{\text{inc}}\)(\%) \\
      \midrule
      \xmark & \xmark & 46.8$\pm$1.2 & 60.9$\pm$1.0 & 37.9$\pm$1.0 & 54.4$\pm$0.8 \\
      \cmark & \xmark & 49.4$\pm$1.0 & 62.1$\pm$1.1 & 40.2$\pm$1.1 & 55.8$\pm$1.0 \\
      \xmark & \cmark & 47.8$\pm$1.1 & 61.8$\pm$1.0 & 39.0$\pm$1.1 & 54.9$\pm$0.9 \\
      % \rowcolor{black!5}%
      \cmark & \cmark & \textbf{50.6$\pm$0.9} & \textbf{63.2$\pm$1.3} &
                         \textbf{41.5$\pm$1.1} & \textbf{56.5$\pm$1.3} \\
      \bottomrule
    \end{tabular}%
  }
  \vspace{-6pt}
\end{wraptable}

\begin{figure}[t]
  \centering
  \begin{subfigure}[t]{0.32\linewidth}
    \centering
    \includegraphics[width=\linewidth]{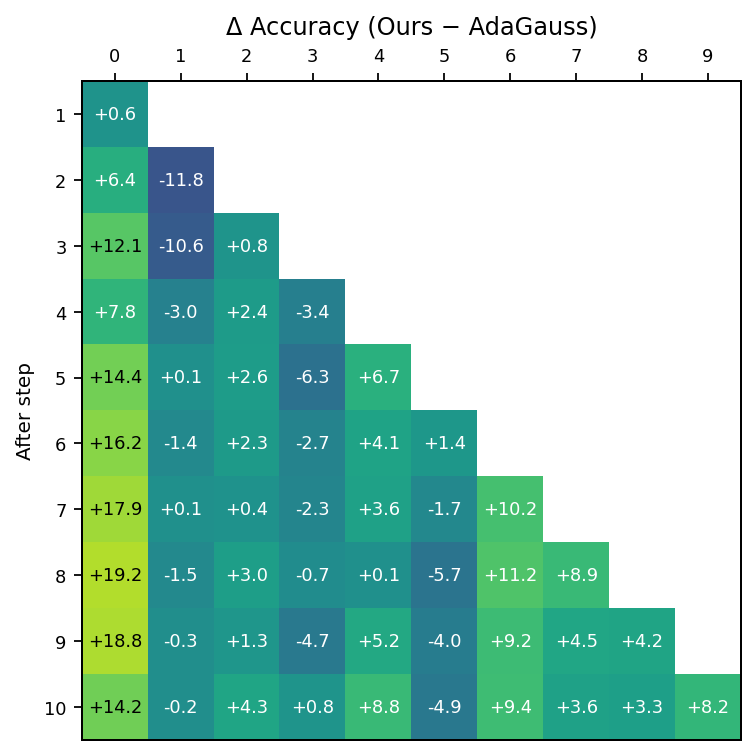}
    % \caption{mu L2}
    \label{fig:three-a}
  \end{subfigure}\hfill
  \begin{subfigure}[t]{0.32\linewidth}
    \centering
    \includegraphics[width=\linewidth]{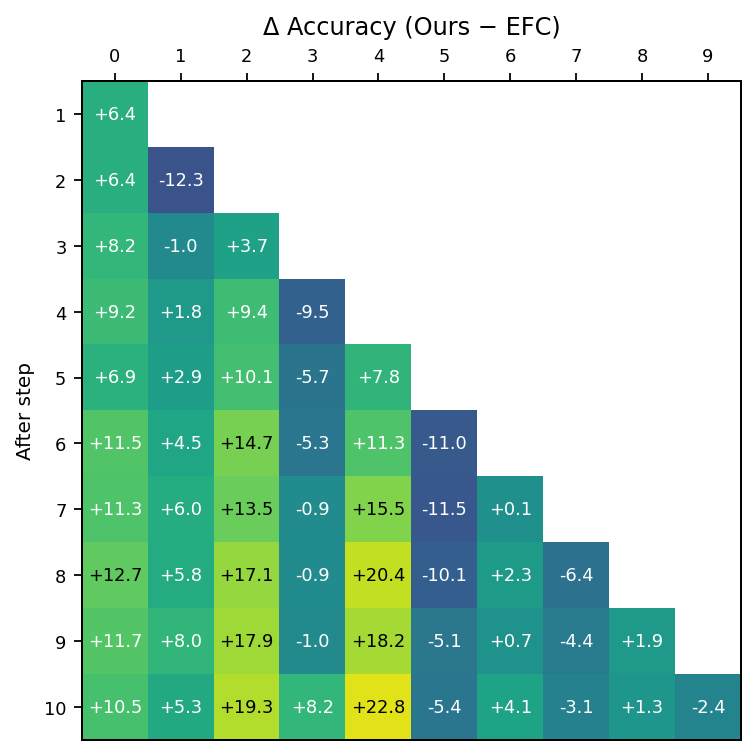}
    % \caption{Cov Frobenius)}
    \label{fig:three-b}
  \end{subfigure}\hfill
  \begin{subfigure}[t]{0.32\linewidth}
    \centering
    \includegraphics[width=\linewidth]{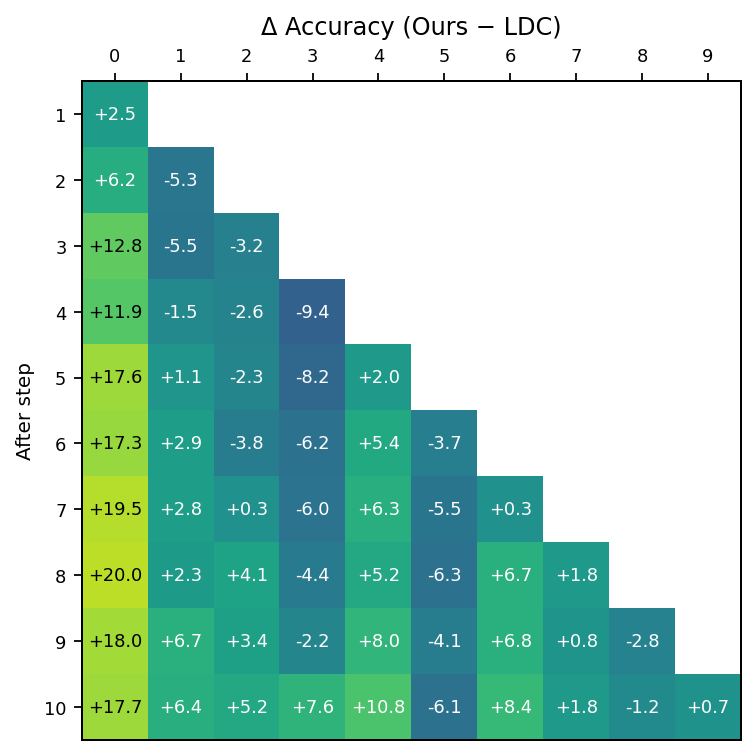}
    % \caption{Symmetric KL }
    \label{fig:three-c}
  \end{subfigure}
  \vspace{-3mm}
  \caption{CIFAR-100 ($T{=}10$): Per-step, per-task accuracy gains ($\Delta$, percentage points) of \textbf{Ours} over AdaGauss, EFC, and LDC. Improvements concentrate on earlier tasks, indicating stronger retention and reduced forgetting.}
  \label{fig:heatmap}
  \vspace{-4mm}
\end{figure}

\vspace{-3mm}
\paragraph{Per-step advantage on CIFAR-100 ($T{=}10$).}
As shown in Figure~\ref{fig:heatmap}, across three baselines, our method shows consistently
positive accuracy gain throughout training, with the \emph{largest gains on
older tasks} (lower-right region in each heatmap). Against EFC, margins
often exceed \textbf{+15--20} pp at mid/late steps; versus LDC, we
sustain \textbf{+6--11} pp on most old tasks; and relative to
AdaGauss we obtain \textbf{+5--10} pp improvements that persist to the
final step. The concentration of positive $\Delta$ on early tasks indicates
\textbf{significantly smaller forgetting}: accuracy on initial tasks decays far
less under ours while recent tasks remain competitive, yielding a superior
plasticity--stability trade-off.

\vspace{-1mm}
\subsection{Advance in Forgetting}
\vspace{-3mm}
\noindent

As shown in Table~\ref{tab:main_results3}, across the three \textbf{balanced, training-from-scratch} datasets, our method achieves the \textbf{lowest forgetting}. On \textbf{CUB-200}, however, most methods fine-tune from a \textbf{pretrained backbone}, so the gaps in forgetting are much smaller than in the from-scratch regime. 
\vspace{-3mm}
\subsection{Effect of \texorpdfstring{$\mathcal{L}_{\text{bi}}$}{L\_bi} and 
\texorpdfstring{$\mathcal{L}_{\text{cyc}}$}{L\_cyc}}
\vspace{-4mm}
\label{sec:ablation_bi_cyc}
\noindent

Notably, our approach delivers \textbf{especially strong preservation of prior knowledge} when training from scratch.

\begin{wrapfigure}[14]{r}{0.37\columnwidth}
  \vspace{-0.3cm}
  \centering
  % \captionsetup{type=figure,font=\scriptsize,labelfont=bf,labelsep=period}
  \includegraphics[width=\linewidth]{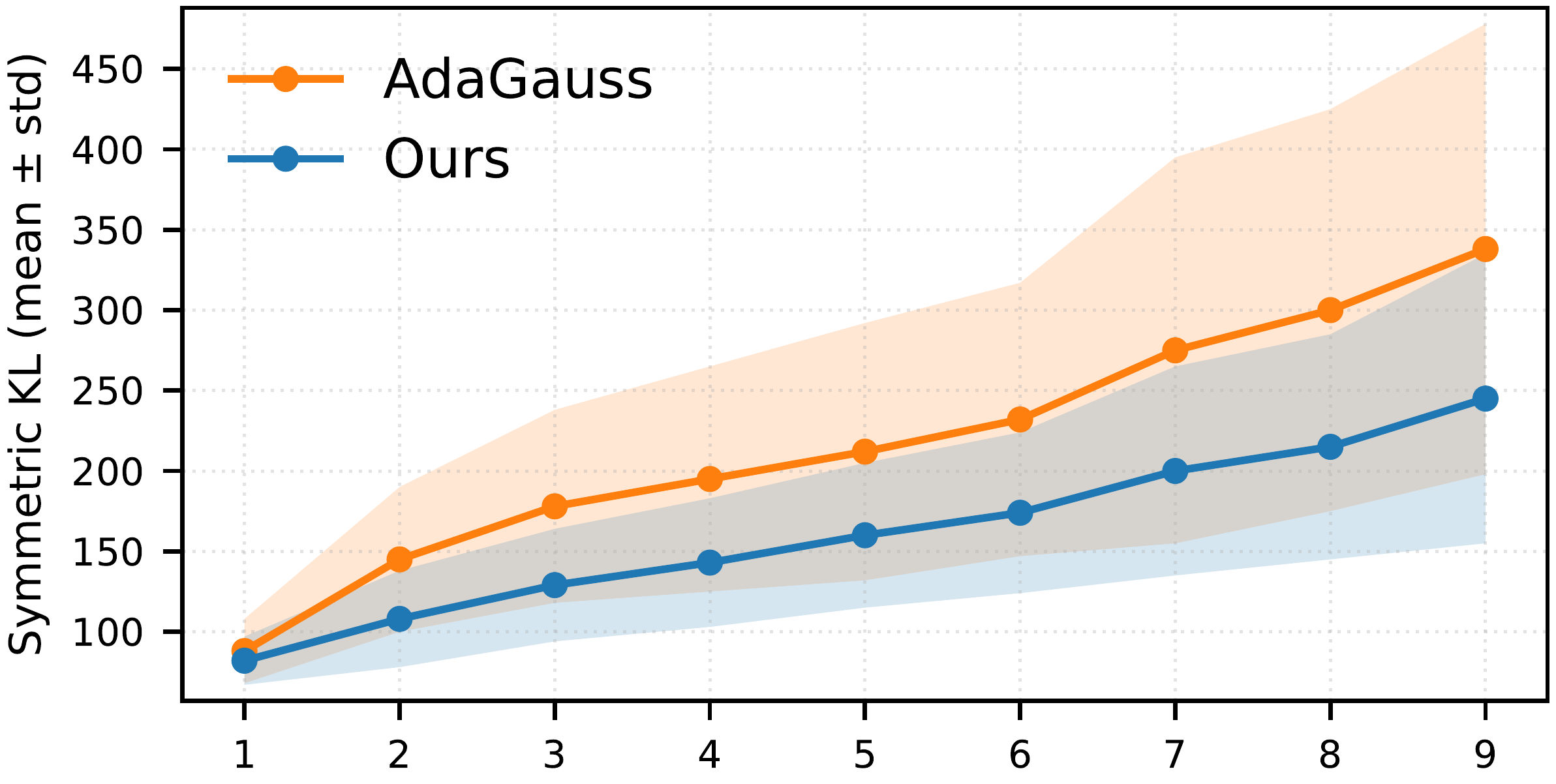}
  \caption{\textbf{Task-0 stability via SymKL (↓).}
  On the fixed task-0 data, we compare Gaussian fits from models after $t{=}1\ldots9$ to the task-0 reference using symmetric KL (Eqs.~\ref{eq:kl-one-way}–\ref{eq:sym-kl}); mean$\pm$std over classes. Our method maintains a smaller divergence—i.e., a closer match to the original distribution—than AdaGauss.}
  \label{fig:symkl_wrap}
  \vspace{-6pt}
\end{wrapfigure}

\begin{wrapfigure}[13]{r}{0.37\columnwidth}
  \vspace{-0.3cm}
  \centering
  % \captionsetup{type=figure,font=\scriptsize,labelfont=bf,labelsep=period}
  \includegraphics[width=\linewidth]{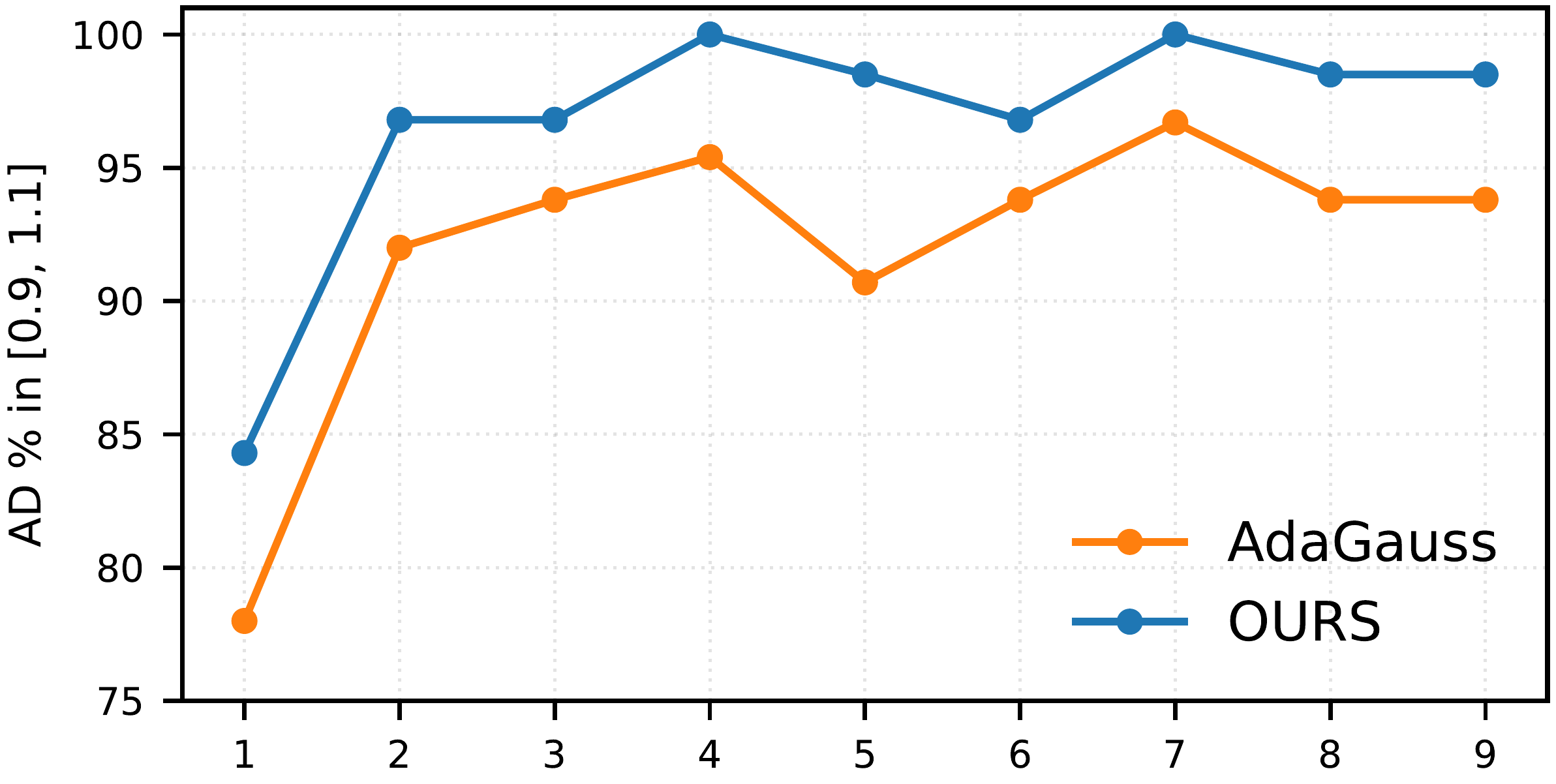}
  \caption{\textbf{Near-isometry on task-0 under continual updates.}
  AD-\% in $[0.9,1.1]$ for models after $t{=}1\ldots9$; our method consistently preserves geometry better than AdaGauss.}
  \label{fig:adsvd_wrap}
  \vspace{-6pt}
\end{wrapfigure}

As summarized in Table~\ref{tab:model_component}, on CIFAR-100 enabling either loss improves both $A_{\text{last}}$ and $A_{\text{inc}}$ over the AdaGauss baseline, and enabling both yields the best results across the 10- and 20-task splits. This pattern matches the roles established in Sec.~\ref{Method}: $\mathcal{L}_{\text{bi}}$ (Eq.~\ref{eq:lbi}) reduces the new$\!\leftrightarrow\!$old feature-transport errors that bound prototype mean/covariance mismatch, while $\mathcal{L}_{\text{cyc}}$ (Eq.~\ref{eq:lcyc}) contracts the spectrum of $AD$ toward $1$, mitigating rank loss and promoting near-isometric transport. Used together, they simultaneously lower transport error and preserve geometry, explaining consistent gains in $A_{\text{last}}$ and $A_{\text{inc}}$. Empirical diagnostics corroborate this: Figs.~\ref{fig:symkl_wrap} and \ref{fig:adsvd_wrap} (CIFAR-100, $T{=}10$) show lower symmetric KL between transported and ground-truth class Gaussians and singular-value spectra of $AD$ that are tighter and more concentrated at $1$ than AdaGauss, indicating better distributional transport and more stable decision boundaries.

\begin{table*}[t]
\centering
\setlength{\tabcolsep}{3pt}
\scriptsize
\caption{Adapter Strategy vs.\ Architecture: Ablation on CIFAR-100}

\begin{subtable}[t]{0.48\textwidth}
\centering
\caption{Direct prototype projection vs.\ projection with post-training adapter fine-tuning. Arrows indicate the preferred direction.}
\label{tab:ablation_cifar100_t10_t20_iclr_small_strategy}
\begin{tabular}{lcccc}
\toprule
\textbf{Ablation} & \multicolumn{2}{c}{$T{=}10$} & \multicolumn{2}{c}{$T{=}20$} \\
\cmidrule(lr){2-3}\cmidrule(lr){4-5}
& $A_{\text{last}}\uparrow$ & $F_{\text{last}}\downarrow$
& $A_{\text{last}}\uparrow$ & $F_{\text{last}}\downarrow$ \\
\midrule
Direct projection              & 49.9 & 15.2 & 38.9 & 17.6 \\
\,+ fine tuning (vs.\ Direct)  & +0.7 & -1.7 & +2.6 & -1.0 \\
\bottomrule
\end{tabular}
\end{subtable}
\hfill
\begin{subtable}[t]{0.48\textwidth}
\centering
\caption{Adapter/Distiller Architectures: MLP shows absolute scores, others report $\Delta$ vs. MLP}
\label{tab:ablation_cifar100_t10_t20_iclr_small_arch}
\begin{tabular}{lcccc}
\toprule
\textbf{Ablation} & \multicolumn{2}{c}{$T{=}10$} & \multicolumn{2}{c}{$T{=}20$} \\
\cmidrule(lr){2-3}\cmidrule(lr){4-5}
& $A_{\text{last}}\uparrow$ & $F_{\text{last}}\downarrow$
& $A_{\text{last}}\uparrow$ & $F_{\text{last}}\downarrow$ \\
\midrule
MLP           & 50.6 & 13.5 & 41.5 & 16.6 \\
Linear & -3.5 & +1.2 & -4.2 & +1.3 \\
CrossAttention& -2.7 & -0.7 & -6.0 & -1.0 \\
MoE& -3.7 & -3.6 & -2.8 & -4.9 \\
\bottomrule
\end{tabular}
\end{subtable}

\vspace{-12pt}
\end{table*}

\vspace{-2mm}
\subsection{Ablation: direct projection vs.\ post-training fine-tuning.}
\vspace{-1mm}
The adapter learned via bidirectional cycle consistency can be used \emph{as is} to map old-class prototypes into the new space. We compare this “Direct projection” with an additional \emph{post-training} fine-tuning of the adapter. On CIFAR-100, direct projection achieves $A_{\text{last}}{=}49.9$ and $F_{\text{last}}{=}15.2$ at 10-task split, and $A_{\text{last}}{=}38.9$ and $F_{\text{last}}{=}17.6$ at 20-task split. Fine-tuning yields consistent gains: $+0.7$ points in $A_{\text{last}}$ and $-1.7$ in $F_{\text{last}}$ at 10-task split, and a larger $+2.6$ / $-1.0$ at 20-task split. These results indicate that while the cycle-consistent adapter already provides a strong zero-shot projection, a brief post-training adjustment further aligns prototypes to the new feature geometry—an effect that becomes more pronounced as the task sequence lengthens.

\vspace{-1mm}
\subsection{Ablation: Adapter/Distiller Architecture}
\vspace{-1mm}
Because our method learns bidirectional maps between old and new feature spaces, the adapter/distiller architecture directly affects performance. Beyond the linear or shallow MLP adapters common in prior work, we test lightweight but richer alternatives—cross-attention and sparse MoE—to probe whether conditional/nonlinear mappings better track representation drift. Table~\ref{tab:ablation_cifar100_t10_t20_iclr_small_arch} reports CIFAR-100 results for the 10- and 20-task splits. Across both splits, multilayer adapters consistently outperform a single linear map: relative to an MLP baseline, the linear variant lowers $A_{\text{last}}$ by $3.5$–$4.2$ points and increases $F_{\text{last}}$ by $1.2$–$1.3$ points. Within the multilayer family, cross-attention favors stability, reducing forgetting ($\Delta F_{\text{last}}=-0.7$ to $-1.0$) at the expense of accuracy ($\Delta A_{\text{last}}=-2.7$ to $-6.0$), whereas sparse MoE delivers the largest forgetting gains ($-3.6$ to $-4.9$) with only moderate accuracy drops ($-2.8$ to $-3.7$). If new and old features differed by a single global affine transform, a linear adapter would suffice; the observed trade-offs instead point to content-dependent, anisotropic drift, which conditional/nonlinear adapters model more faithfully. All variants share identical training schedules; a parameter-matched linear control is a natural follow-up to isolate capacity from architecture.

% \begin{figure}[t]
%   \centering
%   \begin{subfigure}[t]{0.48\linewidth}
%     \centering
%     \includegraphics[width=\linewidth]{figs/Final_drift_by_task_ICLR.png}
%     \caption{}
%     \label{fig:drift-task}
%   \end{subfigure}\hfill
%   \begin{subfigure}[t]{0.48\linewidth}
%     \centering
%     \includegraphics[width=\linewidth]{figs/Final_drift_by_class_ICLR.png}
%     \caption{}
%     \label{fig:drift-class}
%   \end{subfigure}\hfill
%   \caption{CIFAR-100 ($T{=}10$). Drift between maintained prototypes and oracle prototypes (empirical class means) after completing Task~9. For each of the $90$ old classes (Tasks~0--8), we measure the $\ell_2$ distance in feature space between the maintained prototype and its oracle prototype. (a) Per-source-task average drift for the three methods. (b) Histogram of per-class drift over all old classes.}
%   \label{fig:drift-overview}
%   \vspace{-6mm}
% \end{figure}

% % in preamble (if not already):
% \usepackage{wrapfig}

\begin{wrapfigure}{r}{0.55\linewidth}
  \vspace{-2mm}
  \centering
  \begin{minipage}{\linewidth}
    \centering
    \begin{subfigure}[t]{0.49\linewidth}
      \centering
      \includegraphics[width=\linewidth]{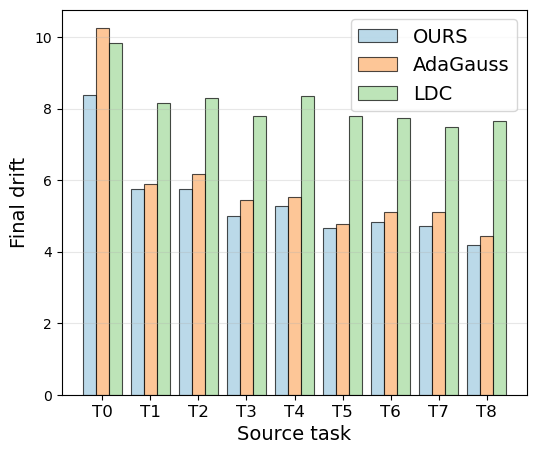}
      \caption{}
      \label{fig:drift-task}
    \end{subfigure}\hfill
    \begin{subfigure}[t]{0.49\linewidth}
      \centering
      \includegraphics[width=\linewidth]{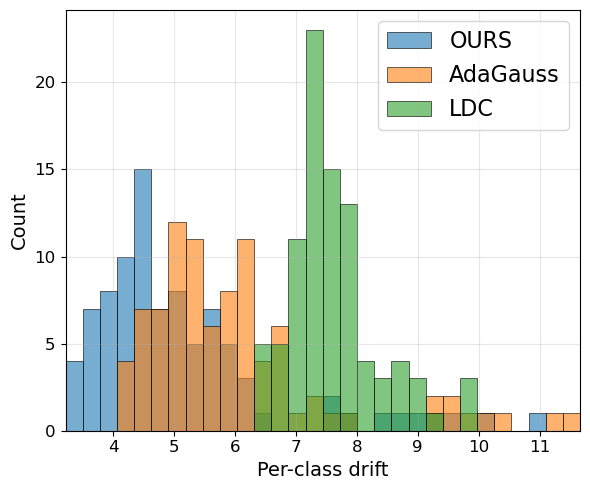}
      \caption{}
      \label{fig:drift-class}
    \end{subfigure}
  \end{minipage}

  \caption{CIFAR-100 ($T{=}10$). Drift between maintained prototypes and oracle prototypes (empirical class means) after completing Task~9. For each of the $90$ old classes (Tasks~0--8), we measure the $\ell_2$ distance in feature space between the maintained prototype and its oracle prototype. (a) Per-source-task average drift for the three methods. (b) Histogram of per-class drift over all old classes.}
  \label{fig:drift-overview}
  \vspace{-4mm}
\end{wrapfigure}

\subsection{Prototype drift from oracle means on CIFAR-100}
\label{sec:drift_oracle}

To assess how well each method preserves old-class geometry, 
Fig.~\ref{fig:drift-overview} reports prototype drift on CIFAR-100 with the
$10$-task split. After training Task~9, we freeze the backbone and, for every
old class $c$, compute the maintained prototype $\widehat{\mu}_c$ and an
\emph{oracle} prototype $\mu_c^\star$ given by the empirical feature mean of
all samples of class $c$ under the final backbone. The drift for class $c$ is
defined as $\lVert \widehat{\mu}_c - \mu_c^\star \rVert_2$.

Panel~\ref{fig:drift-task} averages this drift over the ten classes of each
source task, while Figure~\ref{fig:drift-class} plots the full per-class
distribution over all $90$ old classes. Our method yields both lower average
drift and a tighter distribution at small values than AdaGauss and LDC,
indicating less accumulated distortion of old-class prototypes.

\subsection{Parameter overhead in the 64-dimensional setting}
\label{sec:overhead_ada64}

\textbf{Setup.}
Following AdaGauss~\citep{rypesc2024task}, all experiments use a ResNet-18
backbone followed by a $512{\to}64$ linear reduction and a two-layer MLP
projector $D$ (new$\to$old) in the $S{=}64$ space.
Our bidirectional variant simply adds a second MLP $A$ (old$\to$new) with the
\emph{same} architecture.  
Both $A$ and $D$ are MLPs $\mathbb{R}^S\!\to\!\mathbb{R}^{mS}\!\to\!\mathbb{R}^S$
with width multiplier $m{=}32$ (hidden size $mS{=}2048$).

\textbf{Parameter count.}
A two-layer MLP with biases in this setting has
\[
\#\mathrm{params}_{\text{MLP}}
= 2mS^2 + (m{+}1)S
\quad\Rightarrow\quad
\#\mathrm{params}_{\text{MLP}} = 264{,}256
\]
for $S{=}64$, $m{=}32$.
Thus AdaGauss already uses one such projector $D$ ($\approx\!0.26$M parameters),
and our bidirectional version adds \emph{one more} ($A$), for an extra
\[
\Delta\#\mathrm{params} = 264{,}256
\]
on top of the published AdaGauss model.
Since a standard ResNet-18 backbone has about $11$M parameters, the additional
adapter increases the total parameter count by \textbf{only $\approx 2.4\%$}. We use this shared 64-dimensional configuration in all experiments.
% (We use this shared 64-dimensional configuration in all experiments and please refer
% to Sec.~\ref{sec:adagauss_covariance} and Sec.\ref{sec:overhead_ada64} for a more comprehensive explanation.)

\vspace{-3mm}
\section{Conclusions, Limitations, and Future Work}
\vspace{-3mm}
\label{conclusion}

\noindent \textbf{Conclusions.} We presented a bidirectional drift-compensation framework for exemplar-free class-incremental learning that jointly learns old to new and new to old projectors with stop-gradient gating and cycle consistency. Our analysis links least-squares projectors to CCA and shows how reducing alignment and cycle error stabilizes prototype margins. Experiments across standard EFCIL benchmarks demonstrate strong forgetting reduction, especially in from-scratch settings, while maintaining competitive new-task accuracy.

\noindent \textbf{Limitations.} The current formulation assumes centered features, and second-order (Gaussian) prototype statistics; its theory is local to small alignment errors on the data support. The method may be sensitive to covariance estimation and hyperparameters in low-data regimes.

\noindent \textbf{Future work.} We plan to develop uncertainty-aware and class-imbalance–robust prototype transport, and derive non-asymptotic generalization/forgetting bounds beyond Gaussian assumptions. We also plan to integrate test-time adaptation and multi-modal backbones under strict memory budgets.

\bibliography{iclr2026_conference}
\bibliographystyle{iclr2026_conference}

\appendix

% =========================
% Appendix (camera-ready)
% =========================
\section{Appendix}
\label{sec:appendix}

This appendix provides additional details, proofs, metrics, and supplementary results.
It is organized as follows:
\begin{itemize}
    \item \textbf{Method details.}
    \S\ref{sec:drift_transport_view} (transport view of drift compensation),
    \S\ref{sec:bayes_classifier} (Bayes classifier),
    \S\ref{sec:pseudo} (pseudo-code).
    \item \textbf{Theory.}
    \S\ref{sec:proof_theorem1} (Proof of Theorem~1),
    \S\ref{sec:proof_corollary2} (Proof of Corollary~2).
    \item \textbf{Experimental protocol.}
    \S\ref{sec:setup} (setup),
    \S\ref{sec:acc_metrics} (accuracy metrics),
    \S\ref{sec:dis_metrics} (distribution-similarity metrics),
    \S\ref{sec:ad_band_only} (near-isometry metric).
    \item \textbf{Additional results and ablations.}
    \S\ref{sec:add_visual} (additional visualizations),
    \S\ref{sec:detail_arc} (distiller/adapter architecture ablations),
    \S\ref{sec:param_sensitivity} (hyperparameter sensitivity),
    \S\ref{sec:choice_classifier} (choice of classifier).
    \item \textbf{Additional discussion and diagnostics.}
    \S\ref{sec:proto_prior_work} (prototype-based EFCIL and Gaussian modeling in prior work),
    \S\ref{sec:gaussian_diag} (Gaussianity diagnostics),
    \S\ref{sec:tsne_drift} (t-SNE snapshots),
    \S\ref{sec:intuitive_cycle} (intuitive view of bidirectional cycle consistency),
    \S\ref{sec:adagauss_covariance} (AdaGauss, full-covariance prototypes, and robustness).
    \item \textbf{Limitations and disclosure.}
    \S\ref{sec:1k_limit} (ImageNet-1K limitations),
    \S\ref{sec:llm} (LLM usage disclosure).
\end{itemize}

% ---------------------------------------------------------
% Method details
% ---------------------------------------------------------
\subsection{Method Details}

\subsubsection{Prototype-Drift Compensation: A Transport Perspective}
\label{sec:drift_transport_view}

In the main paper, we adopt a vectorial notion of prototype drift. For each previously-seen class
\(c\in\mathcal{C}_{1:t-1}\), the backbone update from \(f_{t-1}\) to \(f_t\) induces the
feature-mean displacement
\begin{equation}
\label{eq:vec_drift}
\bm{\Delta}_c^{\,t} \;=\; \bm{\mu}_c^{\,t} - \bm{\mu}_c^{\,t-1},
\qquad
\delta_c^{\,t} \;=\; \bigl\|\bm{\Delta}_c^{\,t}\bigr\|_2,
\end{equation}
where \(\bm{\mu}_c^{\,t}\!=\!\frac{1}{|\mathcal{D}_c|}\sum_{x\in\mathcal{D}_c} f_t(x)\)
is the (unknown) class mean under the updated encoder \(f_t\).
Because EFCIL forbids storing past raw samples, \(\bm{\mu}_c^{\,t}\) cannot be recomputed
exactly, and cached prototypes \(\bm{\mu}_c^{\,t-1}\) become stale once \(f_t\) is deployed.

Most drift-compensation pipelines in prior work follow a two-stage pattern: Stage~I constrains the backbone update using only
\(\mathcal{D}_t\),
\begin{equation}
\label{eq:stage1_reg}
\mathcal{L}_{\text{S1}}
=\mathbb{E}_{(x,y)\in\mathcal{D}_t}
\Big[\mathrm{CE}\bigl(g(f_t(x)),y\bigr)
+\lambda\,D\bigl(\phi_{\text{new}}(x),\phi_{\text{old}}(x)\bigr)\Big],
\end{equation}
with \(\phi\in\{f(\cdot),\,g\!\circ\! f(\cdot)\}\) and \(D\) a generic distillation/regularizer.
Stage~II then learns a forward adapter \(A_t\) (with frozen \(f_{t-1},f_t\)) by aligning paired features
\(\bigl(f_{t-1}(x),f_t(x)\bigr)\) on \(\mathcal{D}_t\), and transports old prototypes:
\begin{equation}
\label{eq:transport_update}
A_t \in \arg\min_A \,\mathbb{E}_{x\in\mathcal{D}_t}\!\bigl\|A(f_{t-1}(x)) - f_t(x)\bigr\|_2^2,
\qquad
\tilde{\bm{\mu}}_c^{\,t} \;=\; A_t\!\bigl(\bm{\mu}_c^{\,t-1}\bigr).
\end{equation}
This transport view unifies existing drift-compensation recipes—each can be seen as instantiating
either a global/class-wise translation \(A_t(z)=z+\widehat{\bm{\Delta}}^{\,t}\) or a learned projector
\(A_t\) applied to cached prototypes.

\paragraph{Transport-based summary of prior methods.}
Below we cast representative EFCIL approaches as special cases of
Eq.~\ref{eq:transport_update}. For consistency, we denote the encoders by \(f_{t-1}\) and \(f_t\)
(some works write \(F_{t-1},F_t\)) and use \(\mathcal{D}_t\) for the current-task data.

\begin{itemize}
  \item \textbf{Semantic Drift Compensation (SDC)}~\citep{0004TLHWCJ020}.
  SDC estimates a \emph{global} shift from new-task samples and uses it as a
  translation adapter:
  \[
  \bar{\bm{\Delta}}^{\,t}
  \;=\;
  \frac{1}{|\mathcal{D}_t|}
  \sum_{x\in\mathcal{D}_t}\!\bigl(f_t(x)-f_{t-1}(x)\bigr),
  \qquad
  A_t(z)=z+\bar{\bm{\Delta}}^{\,t},
  \quad
  \tilde{\bm{\mu}}_c^{\,t}=\bm{\mu}_c^{\,t-1}+\bar{\bm{\Delta}}^{\,t}.
  \]

  \item \textbf{Adversarial Drift Compensation (ADC)}~\citep{GoswamiSLKT024}.
  For each old class \(c\), ADC selects a current-sample \(\hat{x}_c\) that is
  adversarially driven towards the vicinity of \(\bm{\mu}_c^{\,t-1}\) (in the old space),
  and takes the resulting pairwise feature gap as a class-wise translation:
  \[
  \widehat{\bm{\Delta}}_c^{\,t}
  \;=\;
  f_t(\hat{x}_c)-f_{t-1}(\hat{x}_c),
  \qquad
  A_t^{(c)}(z)=z+\widehat{\bm{\Delta}}_c^{\,t},
  \quad
  \tilde{\bm{\mu}}_c^{\,t}=\bm{\mu}_c^{\,t-1}+\widehat{\bm{\Delta}}_c^{\,t}.
  \]

  \item \textbf{Learnable Drift Compensation (LDC)}~\citep{GomezVillaGWBTW24}.
  LDC directly \emph{learns} a projector as the adapter:
  \[
  G_\theta\in\arg\min_G\,
  \mathbb{E}_{x\in\mathcal{D}_t}\!\bigl\|G\!\bigl(f_{t-1}(x)\bigr)-f_t(x)\bigr\|_2^2,
  \qquad
  A_t(z)=G_\theta(z),\quad
  \tilde{\bm{\mu}}_c^{\,t}=G_\theta\!\bigl(\bm{\mu}_c^{\,t-1}\bigr).
  \]
  This captures non-linear, potentially class-dependent deformations.

  \item \textbf{EFC (EFM-weighted transport).}~\citep{Magistri2024EFC}
  EFC computes a weighted average of per-sample shifts using a pseudo-metric
  induced by the Empirical Feature Matrix \(E_{t-1}\) (estimated after task \(t{-}1\)).
  Let \(\|v\|_{E}^2:=v^\top E\,v\). Each \(x_i\!\in\!\mathcal{D}_t\) casts a vote
  for class \(c\) with weight
  \[
  w_{c,i}
  \;=\;
  \exp\!\Big(
  -\tfrac{\|\,f_{t-1}(x_i)-\bm{\mu}_c^{\,t-1}\,\|_{E_{t-1}}^2}{2\sigma^2}
  \Big),
  \]
  yielding the class-wise transport
  \[
  \widehat{\bm{\Delta}}_c^{\,t}
  \;=\;
  \frac{\sum_{x_i\in\mathcal{D}_t} w_{c,i}\,\bigl(f_t(x_i)-f_{t-1}(x_i)\bigr)}
       {\sum_{x_i\in\mathcal{D}_t} w_{c,i}},
  \qquad
  \tilde{\bm{\mu}}_c^{\,t}
  \;=\;
  \bm{\mu}_c^{\,t-1}+\widehat{\bm{\Delta}}_c^{\,t}.
  \]

  \item \textbf{AdaGauss.}
  Like LDC, AdaGauss first learns a forward projector \(G_\theta\) by aligning paired features on \(\mathcal{D}_t\):
  \[
  G_\theta \in \arg\min_G\,
  \mathbb{E}_{x\in\mathcal{D}_t}\!\bigl\|G\!\bigl(f_{t-1}(x)\bigr)-f_t(x)\bigr\|_2^2,
  \qquad
  A_t(z)=G_\theta(z).
  \]
  Unlike LDC—which directly transports old \emph{means} via
  \(\tilde{\bm{\mu}}_c^{\,t}=G_\theta\!\bigl(\bm{\mu}_c^{\,t-1}\bigr)\)—AdaGauss models each old class as a Gaussian and
  transports the \emph{distribution} by Monte Carlo push-forward (see Alg.~\ref{alg:bi_consistency}, Stage~II):
  \[
  u_m \sim \mathcal{N}\!\bigl(\bm{\mu}_c^{\,t-1},\,\Sigma_c^{\,t-1}\bigr),\quad
  v_m = G_\theta(u_m)=A_t(u_m),\qquad m=1,\dots,M,
  \]
  followed by re-estimation in the new space:
  \[
  \tilde{\bm{\mu}}_c^{\,t}=\frac{1}{M}\sum_{m=1}^{M} v_m,\qquad
  \tilde{\Sigma}_c^{\,t}=\frac{1}{M-1}\sum_{m=1}^{M}(v_m-\tilde{\bm{\mu}}_c^{\,t})(v_m-\tilde{\bm{\mu}}_c^{\,t})^\top.
  \]
  When \(G_\theta\) (equivalently \(A_t\)) is affine, this reduces in closed form to pushing moments
  \((\tilde{\bm{\mu}}_c^{\,t},\tilde{\Sigma}_c^{\,t})=(A\bm{\mu}_c^{\,t-1}+b,\;A\Sigma_c^{\,t-1}A^\top)\).
\end{itemize}

\subsubsection{Bayes Classifier in the New Feature Space}
\label{sec:bayes_classifier}
Let $z=f_t(x)\in\mathbb{R}^d$ be the feature of an input $x$ at task $t$ and let each seen class
$c\in\mathcal{C}_{1:t}$ be represented in the \emph{new} space by a Gaussian prototype
$\mathcal{N}(\mu_c,\Sigma_c)$ (means and covariances transported/estimated as in Sec.~\ref{sec:drift_transport_view}).
We define the Gaussian negative log-posterior score, up to constants independent of $c$, as
\begin{equation}
\label{eq:bayes_score}
s_c(x)
\;=\;
\bigl(z-\mu_c\bigr)^\top \Sigma_c^{-1}\bigl(z-\mu_c\bigr)
+\log\det \Sigma_c
-2\log \pi_c,
\end{equation}
where $\pi_c$ is the class prior.
In our implementation, we follow the common Mahalanobis-prototype variant and retain only the quadratic term, while the full Gaussian score above is used for the theoretical stability analysis; the same perturbation argument applies after dropping the prior and log-determinant terms.
The \emph{task-agnostic} prediction (TAg) is
\begin{equation}
\label{eq:bayes_pred_tag}
\hat{y}_{\text{TAg}}(x)\;=\;\arg\min_{c\in\mathcal{C}_{1:t}}\, s_c(x).
\end{equation}
When a task-aware (TAw) report is required, we restrict the argmin to the current task's label set
$\mathcal{C}_t$:
\begin{equation}
\label{eq:bayes_pred_taw}
\hat{y}_{\text{TAw}}(x)\;=\;\arg\min_{c\in\mathcal{C}_t}\, s_c(x).
\end{equation}

\subsubsection{Pseudo-code for Our Algorithm}
\label{sec:pseudo}
Algorithm~\ref{alg:bi_consistency} specifies the end-to-end procedure for each task $t$: it learns the current backbone $f_t$ under classification with bidirectional alignment and cycle consistency (via $A$ and $D$), and updates the class prototypes by transporting stored Gaussians into the current feature space for inference.

\algrenewcommand\algorithmicrequire{\textbf{Inputs:}}
\algrenewcommand\algorithmicensure{\textbf{Outputs:}}
\newcommand{\detach}[1]{#1^{(\mathrm{detach})}} % annotation-only, no grad

\begin{algorithm}[H]
\caption{Bidirectional Cycle Consistency (EFCIL)}
\label{alg:bi_consistency}
\small
\begin{algorithmic}
\Require
Task stream $\{\Dt_t\}_{t=1}^{T}$; old backbone $f_{t-1}$ (frozen);
current backbone $f_t$ (learnable); classifier head $g$; \\
adapter $A:\mathbb{R}^d\!\to\!\mathbb{R}^d$ (old$\to$new);
distiller $D:\mathbb{R}^d\!\to\!\mathbb{R}^d$ (new$\to$old); \\
hyperparameters $\lambda_{\mathrm{bi}},\lambda_{\mathrm{cyc}},\alpha$; learning rates $\eta,\eta_A,\eta_D$; batch size $B$; \\
anti-collapse hyperparameters $\kappa,\beta,\varepsilon$; per-class sample count $M$ for distribution transport.
\Ensure
Trained $f_t$, $A$, $D$ for each $t$; transported \emph{means \& covariances} for inference.
\vspace{2pt}

\State \textbf{Initialization:} Copy $f_t \gets f_{t-1}$; randomly initialize $A,D$; freeze $f_{t-1}$.
\For{$t=1,\dots,T$}
  \Statex \textit{\# Stage I: Joint training on current task $\Dt_t$}
  \While{\textit{not converged}}
      \State Sample minibatch $\{(x,y)\}_{b=1}^{B} \sim \Dt_t$.
      \State $z_{\text{old}} \gets f_{t-1}(x)$ \Comment{no gradient}
      \State $z_{\text{new}} \gets f_t(x)$
      \State $\ell_{\text{new}} \gets g(z_{\text{new}})$
      \State \textbf{Bidirectional alignment:}
      \State $\mathcal{L}_{\mathrm{bi}}
      \gets \|D(z_{\text{new}})-z_{\text{old}}\|_2^2
      + \|A(z_{\text{old}})-\detach{z_{\text{new}}}\|_2^2$
      \State \textbf{Cycle consistency:}
      \State $\mathcal{L}_{\mathrm{cyc}}
      \gets \|A(D(z_{\text{new}}))-\detach{z_{\text{new}}}\|_2^2
      + \|D(A(z_{\text{old}}))-\detach{z_{\text{old}}}\|_2^2$
      \State \textbf{Classification:} $\mathcal{L}_{\mathrm{CE}} \gets \mathrm{CE}(\ell_{\text{new}},y)$
      \State \textbf{Robust anti-collapse on features:}
      \State \hspace{0.8em}$\Sigma \gets \tfrac{1}{B-1}(z_{\text{new}}-\bar z)^\top(z_{\text{new}}-\bar z)$;\;
      $\tilde{\Sigma}\!\gets\!\tfrac{1}{2}(\Sigma\!+\!\Sigma^\top)$;\;
      $\hat{\Sigma}\!\gets\!\tilde{\Sigma}+\kappa\,\tfrac{\mathrm{tr}(\tilde{\Sigma})}{d}I+\varepsilon I$
      \State \hspace{0.8em}$\mathcal{L}_{\mathrm{ac}}^{\mathrm{rob}}
      \gets -\tfrac{1}{d}\sum_{i=1}^{d}\min\!\big(\mathrm{chol}(\hat{\Sigma})_{ii},\beta\big)$
      \State \textbf{Total:}
      $\mathcal{L} \gets \mathcal{L}_{\mathrm{CE}}
      + \lambda_{\mathrm{bi}}\mathcal{L}_{\mathrm{bi}}
      + \lambda_{\mathrm{cyc}}\mathcal{L}_{\mathrm{cyc}}
      + \alpha\,\mathcal{L}_{\mathrm{ac}}^{\mathrm{rob}}$
  \EndWhile

  \Statex \textit{\# Stage II: Distribution transport via sampling + adapter fine-tuning (optional)}
  \State Freeze $f_{t-1}$, $f_t$, $D$; fine-tune $A$ on $\Dt_t$ with a small LR by
  minimizing $\|A(z_{\text{old}})-\detach{z_{\text{new}}}\|_2^2$.
  \For{each old class $c \in \mathcal{C}_{1:t-1}$}
      \State Load stored stats $(\mu_c^{\,t-1}, \Sigma_c^{\,t-1})$.
      \State \textbf{Sample old features:} draw $U=\{u_m\}_{m=1}^{M} \sim \mathcal{N}(\mu_c^{\,t-1}, \Sigma_c^{\,t-1})$.
      \State \textbf{Push-forward to new space:} $V = \{v_m\}_{m=1}^{M}$ with $v_m \gets A(u_m)$.
      \State \textbf{Re-estimate in new space:}
      \State \hspace{0.8em}$\tilde{\mu}_c^{\,t} \gets \frac{1}{M}\sum_{m=1}^{M} v_m$,
      \quad
      $\tilde{\Sigma}_c^{\,t} \gets \frac{1}{M-1}\sum_{m=1}^{M} (v_m-\tilde{\mu}_c^{\,t})(v_m-\tilde{\mu}_c^{\,t})^\top$.
  \EndFor
  \State \textbf{Estimate new-class stats} under $f_t$ from $\Dt_t$:
  $(\mu_c^{\,t}, \Sigma_c^{\,t})$ for all $c\in\mathcal{C}_t$.
  \State Build a new prototype collection using
  $\{(\tilde{\mu}_c^{\,t},\tilde{\Sigma}_c^{\,t})\}_{c\in \mathcal{C}_{1:t-1}}$
  and $\{(\mu_c^{\,t},\Sigma_c^{\,t})\}_{c\in \mathcal{C}_t}$.
  \State \textbf{Store} $\{(\mu_c^{\,t},\Sigma_c^{\,t})\}_{c\in\mathcal{C}_{1:t}}$ for the next task.
\EndFor
\end{algorithmic}
\end{algorithm}

% ---------------------------------------------------------
% Theory
% ---------------------------------------------------------
\subsection{Theory}

\subsubsection{Proof of Theorem 1}
\label{sec:proof_theorem1}

\begin{proof}[Proof of Theorem~1 (Cycle contraction)]
Let $M := \tilde A\tilde D - I$ and note that by definition of whitening,
$\EE[\tilde z_{\text{new}}\tilde z_{\text{new}}^\top] = I$ (features are taken to be centered; otherwise replace $z$ by its centered version).
Then
\begin{equation}
\EE\|M\,\tilde z_{\text{new}}\|_2^2
= \EE\big[\tilde z_{\text{new}}^\top M^\top M \tilde z_{\text{new}}\big]
= \Tr\!\big(M^\top M\,\EE[\tilde z_{\text{new}}\tilde z_{\text{new}}^\top]\big)
= \Tr(M^\top M)
= \|M\|_F^2,
\end{equation}
\noindent which yields the stated identity.

For the consequence, write the singular values of $\tilde A\tilde D$ as $\{\sigma_k\}_{k=1}^d$.
Since $M=\tilde A\tilde D-I$, Weyl’s inequality gives
$\max_k |\sigma_k-1| \le \|M\|_2 \le \|M\|_F$.
Thus minimizing $\mathcal{L}_{\text{cyc}}=\EE\|M\tilde z_{\text{new}}\|_2^2=\|M\|_F^2$ forces $\|M\|_F\!\to 0$,
hence $\sigma_k\!\to 1$ for all $k$.
In particular, when the loss is small, all singular values of $\tilde A\tilde D$ lie in a tight neighborhood of $1$, preventing rank/energy loss and preserving local geometry on the data support.
\end{proof}

\subsubsection{Proof of Corollary 2}
\label{sec:proof_corollary2}

\begin{proof}[Proof of Corollary~2 (Decision stability for classification)]
Fix a class $c$ and abbreviate $\mu=\mu_c^t$, $\Sigma=\Sigma_c^t$, $\tilde\mu=\tilde\mu_c^{\,t}$, $\tilde\Sigma=\tilde\Sigma_c^{\,t}$,
$\Delta\mu:=\tilde\mu-\mu$, $\Delta\Sigma:=\tilde\Sigma-\Sigma$.
The Bayes log-score is
$\ell_c(x)=\log\pi_c - \tfrac12\log\det\Sigma - \tfrac12 (x-\mu)^\top \Sigma^{-1}(x-\mu)$.
A first-order expansion in $(\Delta\mu,\Delta\Sigma)$ gives the perturbation
\begin{dmath}
\tilde\ell_c(x)-\ell_c(x)
= -\tfrac12\,\Tr(\Sigma^{-1}\Delta\Sigma)
+\tfrac12\,(x-\mu)^\top \Sigma^{-1}\Delta\Sigma\,\Sigma^{-1}(x-\mu)
+\Delta\mu^\top \Sigma^{-1}(x-\mu)
\;+\; R_c(x),
\end{dmath}
\noindent where $R_c(x)=O(\|\Delta\Sigma\|_2^2+\|\Delta\mu\|_{\Sigma^{-1}}^2)$ by the identities
$\log\det(\Sigma+\Delta\Sigma)=\log\det\Sigma+\Tr(\Sigma^{-1}\Delta\Sigma)+O(\|\Delta\Sigma\|_2^2)$
and $(\Sigma+\Delta\Sigma)^{-1}=\Sigma^{-1}-\Sigma^{-1}\Delta\Sigma\,\Sigma^{-1}+O(\|\Delta\Sigma\|_2^2)$.

Taking absolute values and applying Cauchy–Schwarz and spectral norm bounds,
\begin{dmath}
\big|\tilde\ell_c(x)-\ell_c(x)\big|
\;\le\;
C_\Sigma^{(1)}\,\|\Delta\Sigma\|_2
+ C_\Sigma^{(2)}\,\|\Delta\Sigma\|_2\,\|x-\mu\|_{\Sigma^{-1}}^2\\
+ \|\Delta\mu\|_{\Sigma^{-1}}\,\|x-\mu\|_{\Sigma^{-1}}
+ O(\|\Delta\Sigma\|_2^2+\|\Delta\mu\|_{\Sigma^{-1}}^2),
\end{dmath}
\noindent for constants $C_\Sigma^{(1)},C_\Sigma^{(2)}$ depending only on $\|\Sigma^{-1}\|_2$ (and dimension via standard inequalities).
For a pair $(i,j)$, the log-odds perturbation satisfies by triangle inequality
\begin{equation}
\big|(\tilde\ell_i-\tilde\ell_j)-(\ell_i-\ell_j)\big|
\;\le\;
C_\mu\big(\|\Delta\mu_i\|_{(\Sigma_i^t)^{-1}}+\|\Delta\mu_j\|_{(\Sigma_j^t)^{-1}}\big)
+ C_\Sigma\big(\|\Delta\Sigma_i\|_2+\|\Delta\Sigma_j\|_2\big)
+ O(\cdot),
\end{equation}
where $C_\mu,C_\Sigma$ absorb bounded factors of $\|x-\mu_c^t\|_{(\Sigma_c^t)^{-1}}$ on the evaluation support. Now set $\delta_c:=\|\tilde\mu_c^{\,t}-\mu_c^t\|_{(\Sigma_c^t)^{-1}}$ and invoke the transport-fidelity bounds used in the corollary,
\begin{equation}
\delta_c \;\lesssim\; \sqrt{\varepsilon_{\text{old}\to\text{new}}^2},
\qquad
\|\tilde\Sigma^{\,t}-\Sigma^t\|_2 \;\lesssim\; C_1\sqrt{\varepsilon_{\text{old}\to\text{new}}^2}+C_2\,\varepsilon_{\text{cyc,new}},
\end{equation}
\noindent to obtain
\begin{equation}
\big|(\tilde\ell_i-\tilde\ell_j)-(\ell_i-\ell_j)\big|
\;\lesssim\;
C_\mu(\delta_i+\delta_j) + C_\Sigma\|\tilde\Sigma^{\,t}-\Sigma^t\|_2.
\end{equation}
If the right-hand side is strictly smaller than the Bayes margin $m_{ij}(x):=|\ell_i(x)-\ell_j(x)|$, then the sign of the log-odds is unchanged and the Bayes decision between $i$ and $j$ at $x$ is preserved, as claimed.
\end{proof}

% ---------------------------------------------------------
% Experimental protocol
% ---------------------------------------------------------
\subsection{Experimental Protocol}

\subsubsection{Experimental Setup}
\label{sec:setup}
We utilize a workstation equipped with an NVIDIA RTX~6000 Ada GPU and a Xeon Gold 6448Y CPU to run all the experiments.

\textbf{Datasets.}
We evaluate our method on four canonical continual-learning benchmarks CIFAR-100, TinyImageNet, ImageNet-100 and CUB-200. Each benchmark is instantiated with multiple class-incremental task splits so that each class/image is assigned to exactly one incremental task; no raw samples from previous tasks are revisited in later tasks; only the granularity of the partition changes. We use the official train/val (or test) partitions supplied with each dataset.
\begin{itemize}
\item \textbf{CIFAR-100} consists of 50,000 training and 10,000 test images of size $32\times32$ drawn from 100 classes.
\item \textbf{Tiny-ImageNet} contains 100,000 training and 10,000 validation images at $64\times64$ resolution spanning 200 classes.
\item \textbf{ImageNet-100} (also referred to as \textbf{ImageNet-Subset}) includes 130,000 training and 5,000 validation images preprocessed to $224\times224$ for 100 classes.
\item \textbf{CUB-200} comprises 11,788 bird photographs—5,994 for training and 5,794 for testing—covering 200 fine-grained species. All images are center-cropped and resized to $224\times224$ to match ImageNet preprocessing.
\end{itemize}

\textbf{Testing.}
All results are reported with a test batch size of 512 and no test-time augmentations.

\subsubsection{Accuracy Metrics}
\label{sec:acc_metrics}
We evaluate continual learning along three complementary axes: (i) aggregate predictive performance on seen tasks, (ii) distributional alignment between stored prototypes and the current test distribution, and (iii) near–isometry of the learned transport between old and new representations. This subsection formalizes the first axis.

We report the \textbf{last-task average accuracy} $A_{\text{last}}$, its running mean \textbf{average incremental accuracy} $A_{\text{inc}}$, and the \textbf{last-task average forgetting} $F_{\text{last}}$. Let $a_i^{(K)}$ denote accuracy on task $i$ after training up to task $K$, and let $|\mathcal{C}_i|$ be the number of classes introduced at step $i$. Then
\begin{equation}
\label{eq:metrics}
A_{\text{last}}=\frac{\sum_{i=1}^{K} |\mathcal{C}_i|\,a_i^{(K)}}{\sum_{i=1}^{K} |\mathcal{C}_i|},
\qquad
A_{\text{inc}}=\frac{1}{K}\sum_{j=1}^{K} A_{\text{last}}^{(j)},
\qquad
F_{\text{last}}=\frac{\sum_{i=1}^{K} |\mathcal{C}_i|\,f_i^{(K)}}{\sum_{i=1}^{K} |\mathcal{C}_i|},
\end{equation}
where $f_i^{(K)}=\bigl[\max_{1\le j\le K} a_i^{(j)}-a_i^{(K)}\bigr]_+$ and $A_{\text{last}}^{(j)}$ is $A_{\text{last}}$ evaluated at step $j$.

\subsubsection{Distribution-Similarity Evaluation Metrics}
\label{sec:dis_metrics}
To study prototype drift, we compare stored Gaussian prototypes to test-time class statistics under the current backbone. Let $f_\theta(\cdot)\in\mathbb{R}^S$ denote the feature map, and for each class $c$ let $(\widehat{\mu}_c,\widehat{\Sigma}_c)$ be the stored prototype. Given a held-out set $\mathcal{D}^{\text{test}}_c$, we compute
\[
\mu_c^\star \;=\; \frac{1}{|\mathcal{D}^{\text{test}}_c|}\sum_{x\in \mathcal{D}^{\text{test}}_c} f_\theta(x),
\qquad
\Sigma_c^\star \;=\; \operatorname{Cov}\!\bigl(\{f_\theta(x): x\in \mathcal{D}^{\text{test}}_c\}\bigr)\in\mathbb{R}^{S\times S}.
\]
For numerical stability, all expressions involving covariances use Tikhonov regularization
$\widetilde{\Sigma} \coloneqq \Sigma + \varepsilon I$ with a small $\varepsilon>0$.

We report three per-class discrepancies that emphasize complementary aspects of drift; lower values are better.

\paragraph{(1) Prototype Mean Drift ($\mu$-L2).}
\begin{equation}
\label{eq:mu-l2}
\mu\text{-}\mathrm{L2}_c
\;=\;
\bigl\lVert \widehat{\mu}_c - \mu_c^\star \bigr\rVert_2 .
\end{equation}

\paragraph{(2) Covariance Drift (Frobenius).}
\begin{equation}
\label{eq:sigma-fro}
\Sigma\text{-}\mathrm{F}_c
\;=\;
\bigl\lVert \widehat{\Sigma}_c - \Sigma_c^\star \bigr\rVert_{F}
\;=\;
\sqrt{ \operatorname{tr}\!\Bigl[\bigl(\widehat{\Sigma}_c - \Sigma_c^\star\bigr)^\top
                          \bigl(\widehat{\Sigma}_c - \Sigma_c^\star\bigr)\Bigr] }.
\end{equation}

\paragraph{(3) Symmetric KL Between Gaussians.}
\begin{align}
\label{eq:kl-one-way}
D_{\mathrm{KL}}\!\left(\mathcal{N}(\mu_1,\Sigma_1)\,\Vert\,\mathcal{N}(\mu_2,\Sigma_2)\right)
&= \frac{1}{2}\Bigl[
\operatorname{tr}\!\bigl(\Sigma_2^{-1}\Sigma_1\bigr)
+ (\mu_2-\mu_1)^\top \Sigma_2^{-1} (\mu_2-\mu_1)
- S
+ \ln \frac{\det \Sigma_2}{\det \Sigma_1}
\Bigr],
\end{align}
with $S$ the feature dimension and inverses/determinants taken on regularized covariances.
We report the bi-directional form:
\begin{equation}
\label{eq:sym-kl}
\mathrm{SymKL}_c
\;=\;
D_{\mathrm{KL}}\!\left(\mathcal{N}(\widehat{\mu}_c,\widetilde{\Sigma}_c^{\widehat{}})\,\Vert\,
                       \mathcal{N}(\mu_c^\star,\widetilde{\Sigma}_c^\star)\right)
+
D_{\mathrm{KL}}\!\left(\mathcal{N}(\mu_c^\star,\widetilde{\Sigma}_c^\star)\,\Vert\,
                       \mathcal{N}(\widehat{\mu}_c,\widetilde{\Sigma}_c^{\widehat{}})\right).
\end{equation}

\paragraph{Aggregation over a Class Set.}
For any per-class statistic $m_c\in\{\mu\text{-}\mathrm{L2}_c,\Sigma\text{-}\mathrm{F}_c,\mathrm{SymKL}_c\}$ and class set $\mathcal{C}$ (e.g., a task slice), we report its mean and dispersion:
\begin{equation}
\label{eq:agg}
\overline{m} \;=\; \frac{1}{|\mathcal{C}|} \sum_{c\in\mathcal{C}} m_c,
\qquad
\mathrm{std}(m) \;=\;
\sqrt{\frac{1}{|\mathcal{C}|}\sum_{c\in\mathcal{C}}\bigl(m_c-\overline{m}\bigr)^2}.
\end{equation}

\subsubsection{AD-\% in \texorpdfstring{$[0.9,\,1.1]$}{[0.9,1.1]}}
\label{sec:ad_band_only}
Finally, to probe geometry preservation of the old$\leftrightarrow$new mapping, we measure the fraction of singular values of the composed map that lie in a tight unit band. Consistent with Sec.~\ref{sec:drift_transport_view}, let $f_{t-1}$ and $f_t$ be the frozen previous and current encoders at task $t$, and let $A_t$ (old$\!\to\!$new) and $D_t$ (new$\!\to\!$old) be the learned maps. On a held-out split $\mathcal{V}_t$ restricted to the newly introduced classes $\mathcal{C}_t$, extract paired features
\[
z_{\text{old}}=f_{t-1}(x)\in\mathbb{R}^S,\qquad
z_{\text{new}}=f_t(x)\in\mathbb{R}^S,\qquad x\in\mathcal{V}_t,\; y(x)\in\mathcal{C}_t,
\]
stack them as $Z_{\text{old}},Z_{\text{new}}\in\mathbb{R}^{S\times N}$, and form least-squares surrogates:
\[
\widehat{D}_t=(Z_{\text{old}}Z_{\text{new}}^\top)\,(Z_{\text{new}}Z_{\text{new}}^\top)^{\dagger},\qquad
\widehat{A}_t=(Z_{\text{new}}Z_{\text{old}}^\top)\,(Z_{\text{old}}Z_{\text{old}}^\top)^{\dagger}.
\]
Let $\{\sigma_i\}_{i=1}^S=\sigma(\widehat{A}_t\widehat{D}_t)$ be the singular values. We report
\begin{equation}
\label{eq:ad_band_metric_unified}
\mathrm{AD\mbox{-}\%\;in}\,[0.9,\,1.1]
\;=\;
100 \times \frac{1}{S}\sum_{i=1}^{S}\mathbf{1}\!\left\{\,0.9 \le \sigma_i(\widehat{A}_t\widehat{D}_t) \le 1.1\,\right\}.
\end{equation}
(If $A_t$ or $D_t$ is a single linear layer, its weight can replace the corresponding surrogate.)

\noindent\textbf{Interpretation.}
Higher $\mathrm{AD\mbox{-}\%\;in}\,[0.9,\,1.1]$ indicates that $A_tD_t$ is closer to an isometry with less spectral shrinkage/expansion. This complements Sec.~\ref{sec:dis_metrics}: improved near–isometry typically coincides with lower symmetric KL, indicating better preservation of the class-conditional geometry across tasks.

% ---------------------------------------------------------
% Additional results & ablations
% ---------------------------------------------------------
\subsection{Additional Results and Ablations}

\subsubsection{Additional Visualizations}
\label{sec:add_visual}

\begin{figure}[t]
  \centering
  \begin{subfigure}[t]{0.48\linewidth}
    \centering
    \includegraphics[width=\linewidth]{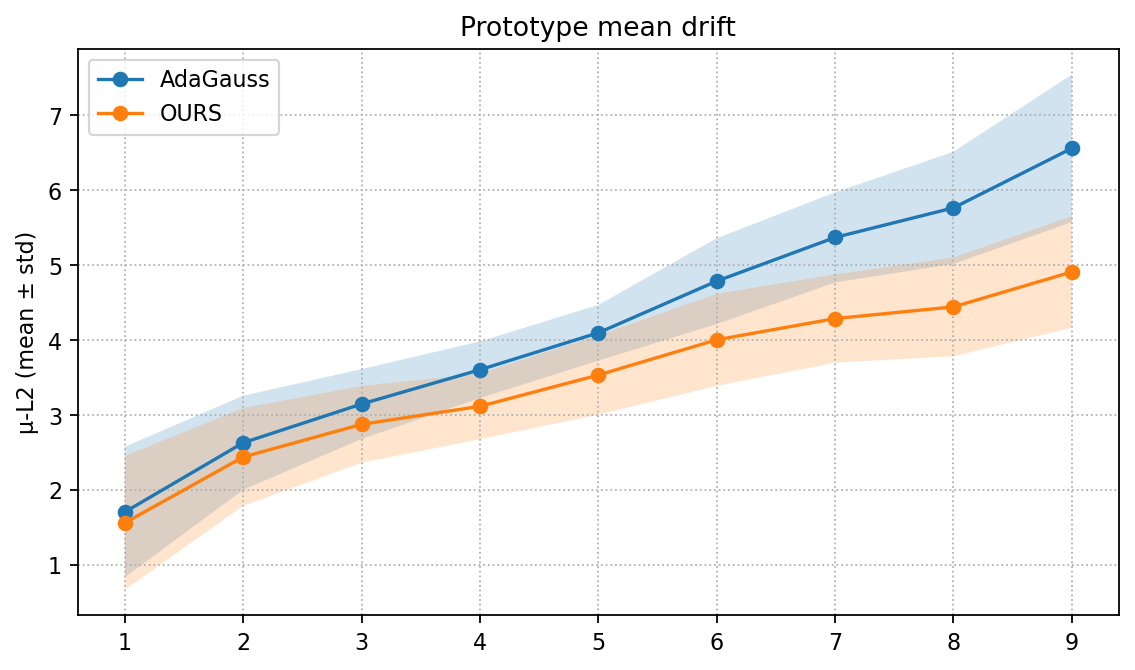}
    \label{fig:sup-drift-a}
  \end{subfigure}\hfill
  \begin{subfigure}[t]{0.48\linewidth}
    \centering
    \includegraphics[width=\linewidth]{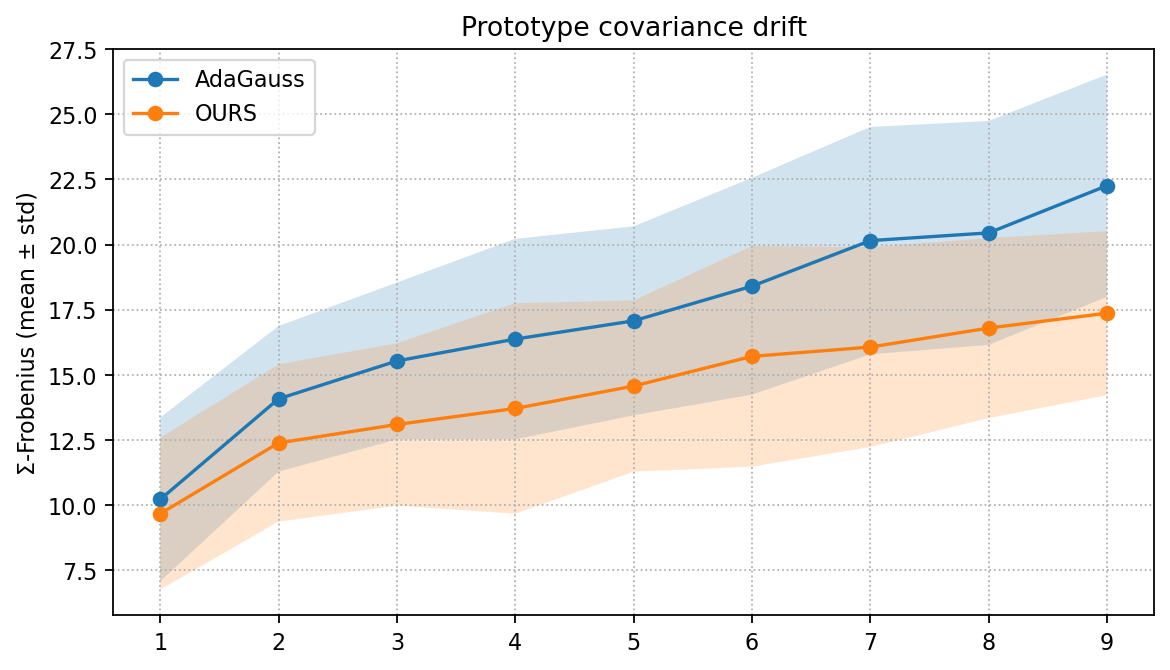}
    \label{fig:sup-drift-b}
  \end{subfigure}\hfill
  \caption{CIFAR-100 ($T{=}10$): \textbf{Prototype drift on task-0 under continual updates (↓).}
Using the fixed task-0 validation split, for each step $t{=}1\ldots9$ we evaluate the model trained up to step $t$.
Left: prototype mean drift $\mu$-L2 (Eq.~\ref{eq:mu-l2}); Right: covariance drift $\Sigma$-Frobenius (Eq.~\ref{eq:sigma-fro}).
Curves show mean$\pm$std over classes (Eq.~\ref{eq:agg}); smaller is better.
\textsc{Ours} exhibits consistently lower center and covariance drift than AdaGauss.}
  \label{fig:three}
\end{figure}

Figure~\ref{fig:three} tracks prototype drift on the fixed task-0 validation split over steps $t{=}1\ldots 9$ (CIFAR-100, $T{=}10$), reporting the mean L2 shift of class centers ($\mu$-L2; Eq.~\ref{eq:mu-l2}) and the Frobenius change of covariances ($\Sigma$-Fro; Eq.~\ref{eq:sigma-fro}) with mean$\pm$std across classes; smaller is better. Our method exhibits consistently lower center and covariance drift than AdaGauss, indicating closer alignment to the original task-0 distribution, i.e., reduced degradation of old-class statistics as $f_t$ evolves.

\noindent\textbf{Relation to the main findings.}
These curves complement the diagnostics in Sec.~\ref{sec:dis_metrics}: we observe both lower symmetric KL between transported and ground-truth Gaussians and a higher fraction of singular values for $A_tD_t$ within $[0.9,1.1]$ (near-isometry), each pointing to better distributional transport and geometry preservation under our bidirectional + cycle training.

\subsubsection{Additional Details on Distiller/Adapter Architecture Ablations}
\label{sec:detail_arc}

\paragraph{Setup and parity.}
All adapter/distiller variants in Table~\ref{tab:ablation_cifar100_t10_t20_iclr_small_arch} are trained under an identical data pipeline, optimization schedule, and loss configuration; only the \emph{architectural family} of the adapter/distiller changes. Each map takes an $S$-dimensional feature and returns an $S$-dimensional output. Unless noted, dropout is disabled and LayerNorms use default $\epsilon$.

\paragraph{Linear.}
A single affine projection $W\!\in\!\mathbb{R}^{S\times S}$ without bias (i.e., $z\mapsto Wz$). This variant is parameter- and compute-light, and serves to illustrate the contribution of our objective under minimal capacity.

\paragraph{MLP (default).}
Unless stated otherwise, we instantiate the adapter/distiller as a \emph{two-layer} MLP
$\mathbb{R}^S \!\to\! \mathbb{R}^{mS} \!\to\! \mathbb{R}^S$
with GELU nonlinearity, no residual connection, and no dropout. We set the width multiplier to $m{=}32$ (hidden size $32S$), which matches the capacity used in our main experiments.

\paragraph{Cross-Attention (XAttn).}
We use a \emph{single} cross-attention block with \emph{pre-LayerNorm}, \emph{8 heads}, and an FFN with \emph{SwiGLU} and expansion $4\times$ (hidden size $4S$), followed by a linear projection back to $S$; dropout is disabled. Queries are produced from current (student) features and keys/values from frozen previous-task (teacher) features, following the standard encoder–decoder attention pattern \citep{vaswani2017attention}.

\paragraph{Mixture-of-Experts (MoE).}
We optionally replace the projection MLP with a \emph{sparse MoE} (Switch-style) comprising \emph{4 experts}. A lightweight router (LayerNorm + linear) performs \emph{top-1} routing per sample; the selected expert is a SwiGLU FFN with expansion $4\times$ (hidden size $4S$) and a linear projection back to $S$; dropout is disabled.

\paragraph{Interpretation and scope.}
Table~\ref{tab:ablation_cifar100_t10_t20_iclr_small_arch} compares representative lightweight instantiations of Linear/MLP/XAttn/MoE under a common training protocol. Since parameter counts and FLOPs naturally co-vary across families, the absolute margins are best read as robustness across common, small-footprint configurations rather than as a strict capacity-matched ranking.

\paragraph{Symbolic capacity accounting (per map).}
Let $S$ denote the feature dimension and $mS$ the MLP hidden size. Ignoring biases, LayerNorm, and constants:
\[
\begin{aligned}
\text{Linear:} \quad & \Theta(S^2) \\
\text{2-layer MLP:} \quad & \Theta(2m\,S^2) \quad (\text{$S\!\to\!mS\!\to\!S$;\ default } m{=}32) \\
\text{1-block XAttn:} \quad & \underbrace{4S^2}_{\text{Q/K/V/O}} \;+\; \underbrace{8S^2}_{\text{FFN }(4\times)} \;\approx\; 12S^2 \\
\text{Sparse MoE (4 experts, top-1):} \quad & \underbrace{O(S^2)}_{\text{router}} \;+\; \underbrace{8S^2}_{\text{active expert per sample}}
\end{aligned}
\]

\begin{table}[h]
  \caption{CIFAR-100: Sensitivity of $\lambda_{\text{bi}}$, $\lambda_{\text{cyc}}$, and $\alpha$.}
  \label{tab:cifar100_ablation}
  \centering
  \small
  \setlength{\tabcolsep}{4pt}
  \renewcommand{\arraystretch}{0.95}
  \resizebox{0.6\linewidth}{!}{%
    \begin{tabular}{@{}ccc|cc|cc@{}}
      \toprule
      \multicolumn{3}{c|}{\textbf{Settings}} &
        \multicolumn{2}{c|}{$T{=}10$} &
        \multicolumn{2}{c}{$T{=}20$} \\
      \cmidrule(lr){1-3}\cmidrule(lr){4-5}\cmidrule(lr){6-7}
      $\lambda_{\text{bi}}$ & $\lambda_{\text{cyc}}$ & $\alpha$ &
      \(A_{\text{last}}\)(\%) & \(A_{\text{inc}}\)(\%) &
      \(A_{\text{last}}\)(\%) & \(A_{\text{inc}}\)(\%) \\
      \midrule
      5   & 1   & 1   & 50.6 & 63.2 & 41.5 & 56.5 \\
      0   & 1   & 1   & 47.8 & 61.8 & 39.0 & 54.9 \\
      5   & 0   & 1   & 49.4 & 62.1 & 40.2 & 55.8 \\
      0   & 0   & 1   & 46.8 & 60.9 & 37.9 & 54.4 \\
      5   & 1   & 0   & 49.7 & 62.3 & 39.2 & 55.2 \\
      5   & 1   & 0.5 & 51.0 & 63.4 & 42.4 & 56.1 \\
      5   & 1   & 2   & 48.7 & 61.9 & 42.6 & 56.5 \\
      0.5 & 1   & 1   & 47.4 & 61.3 & 39.8 & 55.3 \\
      1   & 1   & 1   & 51.3 & \textbf{63.7} & 40.0 & 56.4 \\
      10  & 1   & 1   & 47.2 & 59.9 & 38.8 & 53.8 \\
      5   & 0.5 & 1   & 50.4 & 63.1 & 40.8 & 55.7 \\
      5   & 2   & 1   & \textbf{51.9} & 63.5 & \textbf{42.9} & \textbf{57.0} \\
      \bottomrule
    \end{tabular}%
  }
  \vspace{-6pt}
\end{table}

\subsubsection{Parameter Sensitivity and Choice of Default Hyperparameters}
\label{sec:param_sensitivity}

In the main experiments we did not perform an extensive grid search.
Instead, we chose the scales of the bidirectional and cycle-consistency losses
based on their rough magnitude: $\lambda_{\text{bi}}{=}5$ and
$\lambda_{\text{cyc}}{=}1$ were selected so that the additional terms had a
similar order of contribution as the task loss and KD loss.
For the anti-collapse loss we inherit the default scaling factor
$\alpha{=}1$ from AdaGauss; our robust formulation preserves the role and typical scale of this regularizer.

The parameter-sensitivity study in Table~\ref{tab:cifar100_ablation} varies
$\lambda_{\text{bi}}\in\{0,0.5,1,5,10\}$,
$\lambda_{\text{cyc}}\in\{0,0.5,1,2\}$, and
$\alpha\in\{0,0.5,1,2\}$.
Alternative configurations can slightly outperform our default choice on CIFAR-100.
Nevertheless, for all main results we retain the original defaults to avoid the impression that gains are driven by aggressive hyperparameter tuning.
The bidirectional and cycle-consistency terms improve over the one-directional baseline in most non-degenerate settings, while overly large weights can hurt performance, as expected for strong auxiliary regularization.

\subsubsection{Choice of Classifier}
\label{sec:choice_classifier}

\begin{table}[h]
\vspace{-0.5em}
\centering
\caption{Linear classifier vs.\ Bayesian classifier on CIFAR-100.}
\label{tab:ablation_cifar100_t10_t20_classifier}
\small
\begin{tabular}{lcccc}
\toprule
\textbf{Classifier} & \multicolumn{2}{c}{$T{=}10$} & \multicolumn{2}{c}{$T{=}20$} \\
\cmidrule(lr){2-3}\cmidrule(lr){4-5}
& $A_{\text{last}}\uparrow$ & $A_{\text{inc}}\uparrow$
& $A_{\text{last}}\uparrow$ & $A_{\text{inc}}\uparrow$ \\
\midrule
Bayesian & 50.6 & 63.2 & 41.5 & 56.5 \\
Linear (sampling)   & 51.1 & 63.7 & 40.8 & 55.7 \\
\bottomrule
\end{tabular}
\vspace{-0.5em}
\end{table}

Following common terminology in Gaussian-prototype EFCIL, we refer to our inference rule as the \emph{Bayesian} classifier, although the implementation follows the Mahalanobis-prototype variant described in Sec.~\ref{sec:bayes_classifier}: it retains the quadratic term of the Gaussian score and predicts by distance to stored Gaussian prototypes.
We additionally evaluate a \emph{linear} classifier trained from these Gaussians (Table~\ref{tab:ablation_cifar100_t10_t20_classifier}).
Following the public AdaGauss implementation, we construct a synthetic training set by sampling features from each class-wise Gaussian $\mathcal{N}(\mu_c,\Sigma_c)$ and optimize a single linear head over all seen classes with standard cross-entropy (denoted \emph{Linear (sampling)}).
In the EFC literature~\citep{Magistri2024EFC,magistri2025efcelasticfeatureconsolidation}, this procedure is often referred to as \textbf{Gaussian rebalancing}.

As shown in Table~\ref{tab:ablation_cifar100_t10_t20_classifier}, the sampling-based linear head closely matches the Bayesian classifier: differences in $A_{\text{last}}$ and $A_{\text{inc}}$ remain below one percentage point for both $T{=}10$ and $T{=}20$.
This indicates that our conclusions are not sensitive to the choice between the Bayesian/Mahalanobis-prototype inference rule and a Gaussian-sampling linear head.

% ---------------------------------------------------------
% Additional discussion & diagnostics
% ---------------------------------------------------------
\subsection{Additional Discussion and Diagnostics}

\subsubsection{Prototype-Based EFCIL and Gaussian Modeling in Prior Work}
\label{sec:proto_prior_work}

Prototype-based strategies are a well-established line of work in exemplar-free class-incremental learning (EFCIL). Broadly, existing methods differ in how they represent class prototypes (means vs.\ Gaussians) and how they use them (direct classification vs.\ pseudo-feature rehearsal vs.\ drift compensation).

\textbf{Mean prototypes with synthetic feature rehearsal.}
PASS stores one feature mean per class and performs prototype rehearsal by injecting Gaussian noise around these means to synthesize pseudo-features, which are mixed with current-task data to train the classifier; a self-supervised rotation head is added to further stabilize the backbone.
FeTrIL also stores class means, but produces old-class pseudo-features by a geometric translation of real features from the current task,
\(
\hat f = f_{\text{new}} + \mu_{\text{old}} - \mu_{\text{new}},
\)
and uses these translated features together with new-class features to train a linear classifier.

\textbf{Explicit Gaussian prototypes and covariance-aware classification.}
FeCAM estimates per-class means and covariances and performs Bayes/Mahalanobis classification in this Gaussian space.
EFC represents each class as a Gaussian prototype $(\mu_c,\Sigma_c)$ and samples from these Gaussians to perform asymmetric prototype rehearsal (PR-ACE), while explicitly compensating prototype drift across tasks.
AdaGauss likewise models each class as $\mathcal N(\mu_c,\Sigma_c)$ and introduces an anti-collapse regularizer based on the Cholesky factor of $\Sigma_c$ to prevent rank deficiency and feature collapse, together with covariance-adaptation mechanisms that update $(\mu,\Sigma)$ across tasks.

\textbf{Mean-only drift compensation.}
LDC stores one mean prototype per class and learns a forward projector that maps old-space means into the new feature space after each task.
ADC uses adversarially perturbed current-task inputs whose embeddings lie near old-class means and uses the resulting feature displacements to estimate how old means should move in the new space.
Neither LDC nor ADC explicitly models full Gaussian structure.

\textbf{Our position.}
Our work does not introduce Gaussian prototypes as a new concept; instead, we build on Gaussian-based EFCIL (FeCAM, EFC, AdaGauss) and mean-based drift compensation (LDC, ADC). Our novelty lies in how prototypes are \textbf{transported across tasks}: we introduce a bidirectional projector with cycle consistency that is trained online during backbone optimization, with theoretical guarantees linking the cycle loss to spectral contraction and classification stability.

\subsubsection{Measuring the Adherence of Features to Gaussian Assumptions}
\label{sec:gaussian_diag}

\textbf{On the suitability of multivariate normality tests.}
Mardia's multivariate normality test is not well aligned with the geometry and scale of continual-learning vision features. It relies on third- and fourth-order moments (multivariate skewness and kurtosis), and its asymptotic calibration assumes moderate dimension and i.i.d.\ samples. In high-dimensional settings with complex dependence structures and large sample sizes—as in deep feature spaces of EFCIL benchmarks—this test can become overly restrictive and reject even when deviations are mild and do not affect downstream methods \cite{EbnerHenze2020,ChenXia2023}. Prior analyses also highlight sensitivity to sample size and dimensionality, motivating the use of graphical or geometric diagnostics.

Instead of relying on a global hypothesis test, we assess Gaussianity through a geometric, class-wise visualization in a low-dimensional embedding space. For a fixed subset of classes, we periodically extract their validation features across the training sequence, embed them with t-SNE, and overlay the corresponding fitted class-conditional Gaussians by plotting their one- and two-standard-deviation ellipses. This procedure directly reveals whether the learned representations form compact, approximately elliptical clusters that are stable over time.

\subsubsection{t-SNE Snapshots of Task-0 Classes on CIFAR-100 (10 Tasks)}
\label{sec:tsne_drift}

\begin{figure}[h]
\centering
\begin{subfigure}{0.48\linewidth}
    \centering
    \includegraphics[width=\linewidth]{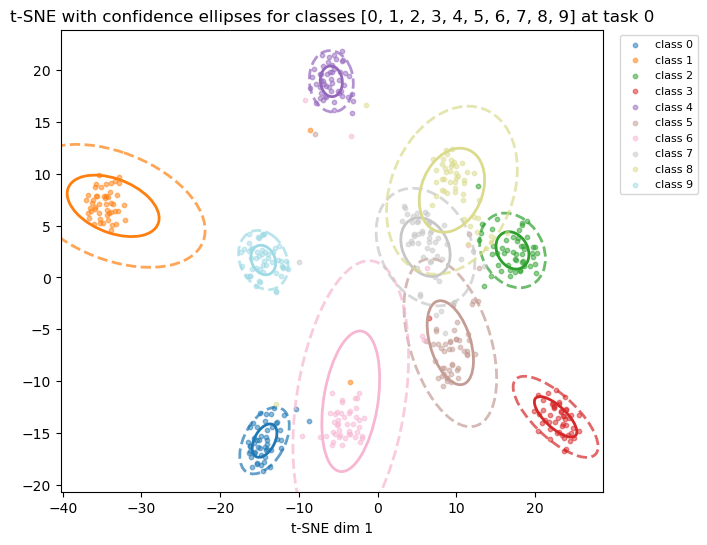}
    \caption{After task 0.}
    \label{fig:tsne:a}
\end{subfigure}
\hfill
\begin{subfigure}{0.48\linewidth}
    \centering
    \includegraphics[width=\linewidth]{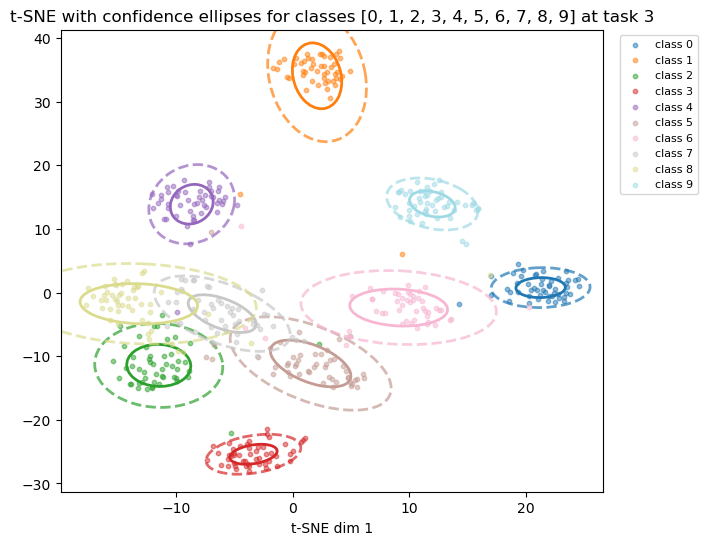}
    \caption{After task 3.}
    \label{fig:tsne:b}
\end{subfigure}

\vspace{0.4em}

\begin{subfigure}{0.48\linewidth}
    \centering
    \includegraphics[width=\linewidth]{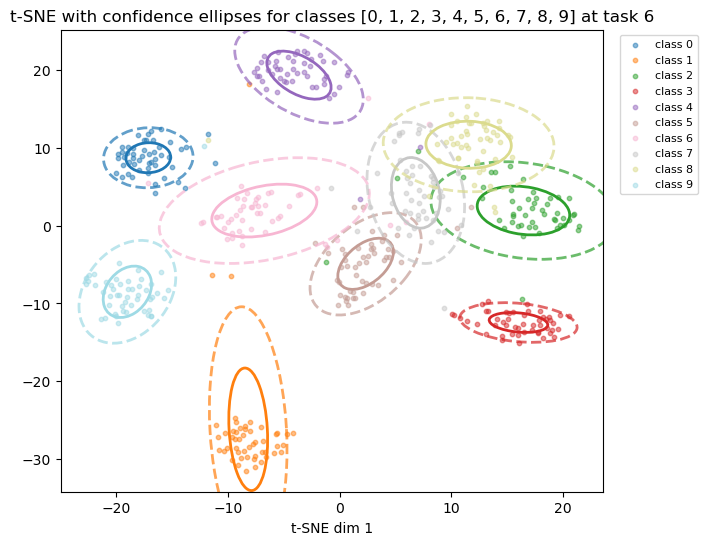}
    \caption{After task 6.}
    \label{fig:tsne:c}
\end{subfigure}
\hfill
\begin{subfigure}{0.48\linewidth}
    \centering
    \includegraphics[width=\linewidth]{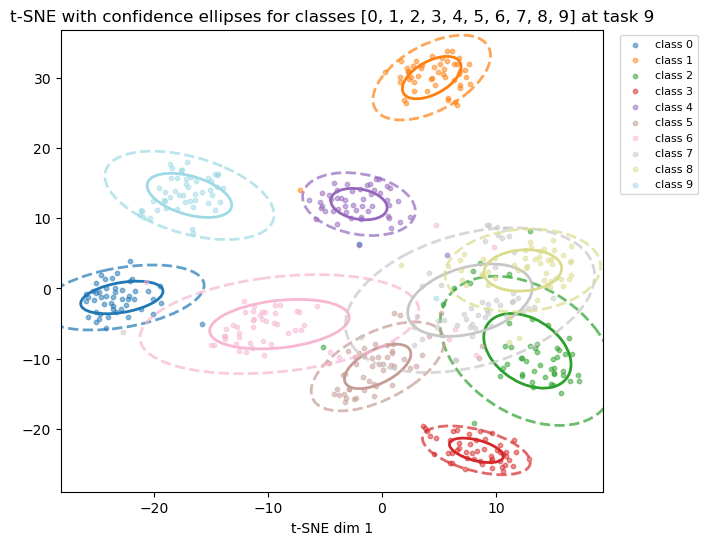}
    \caption{After task 9.}
    \label{fig:tsne:d}
\end{subfigure}

\caption{t-SNE of task-0 classes on CIFAR-100 with $T{=}10$.
We project validation features of the same ten classes after training tasks
0, 3, 6, and 9. Solid and dashed ellipses mark the one- and two-standard-deviation
regions of the fitted Gaussian for each class.}
\label{fig:tsne_drift}
\end{figure}

To better understand how feature distributions evolve over time, we conduct a
t-SNE study on the \emph{balanced} CIFAR-100 benchmark with $T{=}10$ equally
sized tasks. We fix the ten classes introduced at task 0 and, after finishing
tasks 0, 3, 6, and 9, extract their validation features and project them with
t-SNE. For each snapshot in Figure~\ref{fig:tsne_drift}, we fit a Gaussian to
the features of each class and visualize its one- and two-standard-deviation
regions with solid and dashed ellipses, respectively.

Across all stages, the per-class clusters remain roughly unimodal and are well
covered by a single Gaussian, rather than fragmenting into multiple disjoint
modes. This suggests that the dominant challenges in EFCIL are not due to a gross
mismatch of the Gaussian prototype assumption, but rather due to drift of class
statistics and reduced separability as the representation evolves.

\subsubsection{Intuitive View of Bidirectional Cycle Consistency and Low-Drift Regimes}
\label{sec:intuitive_cycle}

\textbf{From post-hoc adapters to in-task bidirectional alignment.}
Most drift-compensation pipelines follow a two-stage pattern: during Stage~I the new encoder $f_t$ is regularized toward $f_{t-1}$ (often via distillation), and only in Stage~II is an adapter $A$ trained post hoc to map old features into the new space. Our goal is to make this duality explicit and move it inside Stage~I: we jointly learn a distiller $D\colon z_{\text{new}}\!\to\!z_{\text{old}}$ and an adapter $A\colon z_{\text{old}}\!\to\!z_{\text{new}}$ while the backbone is still being optimized. Intuitively, $D$ regularizes $f_t$ toward the frozen teacher, while $A$ learns the forward transport used for prototype relocation under the evolving representation.

\textbf{Why this is not adversarial training.}
Cycle consistency in Eq.~\ref{eq:lcyc} is a self-consistency constraint rather than an adversarial game: applying $D$ then $A$ (or $A$ then $D$) should approximately return the original feature on the data support. A key design choice is that both $L_{\text{bi}}$ and $L_{\text{cyc}}$ are implemented with stop-gradient targets: $\|A(z_{\text{old}}) - \text{stopgrad}(z_{\text{new}})\|^2$ updates $A$ only, so $A$ follows the evolving $f_t$ rather than dragging it; the cycle terms update $(A,D)$ but not $f_t$, stabilizing the maps without reducing backbone plasticity.

\textbf{An intuitive reading of Theorem~1 and Corollary~2.}
Theorem~1 analyzes the cycle loss in a whitened feature space where each side has identity covariance; the expected cycle error is the squared Frobenius distance between the composed map $\tilde A \tilde D$ and the identity. Minimizing $L_{\text{cyc}}$ therefore encourages a near-isometry on the data support (singular values close to $1$), preventing rank/energy loss. Corollary~2 then links transport fidelity of means/covariances to bounded perturbations of Bayes scores; if the perturbation is smaller than the margin, old decisions are preserved.

\textbf{Why CUB-200 shows smaller gains.}
On CUB-200 we fine-tune from an ImageNet-pretrained ResNet-18 with a very low backbone learning rate, so representation drift is modest. In this low-drift regime, the incremental benefit of aggressively regularizing and transporting features is naturally smaller; empirically the method behaves closer to AdaGauss.

\subsubsection{On AdaGauss, Full-Covariance Prototypes, and the Need for Robustness}
\label{sec:adagauss_covariance}

\textbf{AdaGauss and full-covariance prototypes.}
AdaGauss represents each class $c$ with a Gaussian prototype $\mathcal{N}(\mu_c, \Sigma_c)$ and uses these Gaussians both for classification (via a Bayes classifier) and for Gaussian sampling to train the prototype adapter. To make this feasible and numerically stable, AdaGauss introduces an anti-collapse loss that regularizes class-wise covariance matrices through a Cholesky factorization.

\textbf{Why dimensionality reduction is necessary.}
The Cholesky-based anti-collapse term and Gaussian sampling require each $\Sigma_c$ to be symmetric positive-definite. In exemplar-free incremental settings, keeping the feature dimension at $512$ (ResNet-18 penultimate layer) often makes reliable full-rank covariance estimation difficult. AdaGauss therefore applies a learned linear reduction layer after the ResNet-18 backbone, mapping $512 \to 64$, improving the sample-to-dimension ratio and stabilizing covariance estimation.

\textbf{Why dimensionality reduction alone is still not sufficient.}
Even after projecting to $S\!=\!64$, the mini-batch covariance can be non-SPD or ill-conditioned in realistic EFCIL regimes (e.g., small $B$, correlated features, or imbalance), leading to Cholesky failures and unstable scales. We therefore adopt a robust anti-collapse variant: symmetrization, shrinkage, and jitter (and fallbacks when needed), aimed at guaranteeing numerical stability without changing the role of the regularizer.

\textbf{Implications for our method.}
Following the public AdaGauss codebase, we retain the 512$\to$64 projection. All Gaussian prototypes, anti-collapse losses, and transport maps are defined in the same $S\!=\!64$ feature space, enabling direct comparisons and controlled ablations.

% ---------------------------------------------------------
% Limitations & disclosure
% ---------------------------------------------------------
\subsection{Limitations and Disclosure}

\subsubsection{Limitations of ImageNet-1K Experiments}
\label{sec:1k_limit}
Under our current setup, scaling this protocol to the full 1K-class ImageNet dataset would require several weeks of continuous GPU time, making such an experiment unrealistic for the present study. Consequently, we restrict large-scale evaluation to ImageNet-100, whose class count still exposes the challenges of our setup while remaining computationally feasible.

\subsubsection{LLM Usage Disclosure}
\label{sec:llm}
We used ChatGPT (OpenAI) as a writing copilot to critique and polish the prose (clarity, tone, and grammar). The model was not used to generate technical content, figures, or results, nor to design experiments or draw conclusions. The authors take full responsibility for all claims and the accuracy of the paper.

\end{document}